%% file: iclr2027_conference.tex
\documentclass{article} 
\usepackage{iclr2027_conference,times}

\usepackage{amsmath,amssymb,amsthm}
\usepackage{bm}
\usepackage{anyfontsize}

\usepackage{graphicx}
\usepackage{xcolor}
\usepackage{array}
\usepackage{booktabs}
\usepackage{multirow}
\usepackage{tabularx}
\usepackage{colortbl}
\usepackage{caption}
\usepackage{subcaption}

\usepackage{algorithm}
\usepackage{algpseudocode}
\usepackage{tikz}
\usetikzlibrary{arrows.meta,positioning,fit,calc,backgrounds}

\usepackage{enumitem}
\usepackage{url}

\input{math_commands.tex}

\usepackage{hyperref}
\usepackage[capitalize,nameinlink]{cleveref}

\definecolor{PQTInk}{HTML}{385A75}
\definecolor{PQTBand}{HTML}{EDF2F6}
\definecolor{PQTOurs}{HTML}{E8F0F6}
\definecolor{PQTOffQ}{HTML}{F8EEE7}
\definecolor{PQTBaseQ}{HTML}{F4F1EB}
\definecolor{PQTRef}{HTML}{F4F3F6}
\definecolor{PQTMuted}{HTML}{64748B}

\definecolor{PQTWarm}{HTML}{98613D}
\definecolor{PQTGreen}{HTML}{426D5A}
\definecolor{PQTMint}{HTML}{EDF4EF}
\definecolor{PQTLilac}{HTML}{70627F}

\definecolor{prism}{HTML}{1D3557}
\definecolor{hadamard}{HTML}{A8DADC}
\definecolor{identity}{HTML}{457B9D}
\definecolor{duquant}{HTML}{E63946}
\definecolor{honeydew}{HTML}{F1FAEE}

\providecommand{\pq}{PrismQuant}

\providecommand{\pqtbest}[1]{\textbf{#1}}

\providecommand{\pqtours}{%
    \textcolor{PQTInk}{\textbf{\pq{}}}%
}

\providecommand{\pqtvenue}[2]{%
    \nobreak\hspace{0.35em}%
    {\fontsize{7pt}{8.3pt}\selectfont
     \textcolor{PQTMuted}{(#1~\textquoteright#2)}}%
}

\providecommand{\pqtdepours}{\pqtours{}}
\providecommand{\pqtdepbest}[1]{\pqtbest{#1}}

\providecommand{\pqtdashline}{%
    \noalign{%
        \vskip 1.5pt
        \nointerlineskip
        \hbox to \linewidth{%
            \color{black}%
            \cleaders\hbox{%
                \vrule width 2pt height 0.4pt depth 0pt
                \kern 1.5pt
            }\hfill
        }%
        \vskip 1.5pt
        \nointerlineskip
    }%
}

\algrenewcommand\algorithmicrequire{%
    \textcolor{PQTInk}{\textbf{Input:}}%
}

\algrenewcommand\algorithmicensure{%
    \textcolor{PQTInk}{\textbf{Output:}}%
}

\algrenewcommand\algorithmiccomment[1]{%
    \hfill\textcolor{PQTMuted}{// #1}%
}

\algrenewcommand\alglinenumber[1]{%
    \scriptsize\textcolor{PQTMuted}{#1}%
}

\providecommand{\PQOp}[1]{\operatorname{#1}}

\providecommand{\PQStage}[3]{%
    \Statex\vspace{2pt}%
    \begingroup
    \setlength{\fboxsep}{3pt}%
    \colorbox{#1}{%
        \parbox{\dimexpr\linewidth-2\fboxsep\relax}{%
            \raggedright
            \strut\textcolor{#2}{\bfseries #3}\strut\par
        }%
    }%
    \endgroup
    \vspace{1pt}%
}

\providecommand{\PQNote}[1]{%
    \hfill{\footnotesize\color{PQTMuted}#1}%
}

\theoremstyle{plain}

\newtheorem{proposition}{Proposition}

\theoremstyle{definition}

\theoremstyle{remark}

\crefname{theorem}{Theorem}{Theorems}
\crefname{lemma}{Lemma}{Lemmas}
\crefname{proposition}{Proposition}{Propositions}
\crefname{corollary}{Corollary}{Corollaries}
\crefname{definition}{Definition}{Definitions}
\crefname{remark}{Remark}{Remarks}

\crefname{section}{Section}{Sections}
\crefname{subsection}{Section}{Sections}
\crefname{subsubsection}{Section}{Sections}
\crefname{appendix}{Appendix}{Appendices}

\crefname{figure}{Figure}{Figures}
\crefname{table}{Table}{Tables}
\crefname{algorithm}{Algorithm}{Algorithms}

\title{PrismQuant: Optimal Null-Space Rotations for Grouped Quantizers}
\title{PrismQuant: Optimal Null-Space Rotations for Grouped Quantizers}

\author{Yanlong Chen \\
Nanyang Technological University \\
Singapore \\
\texttt{yanlong.chen@ntu.edu.sg} \\
\And
Yining Chen \\
Nanjing University of Posts and Telecommunications \\
Nanjing, China \\
\texttt{b24021105@njupt.edu.cn} \\
\And
Song Zhang \\
Nanjing University \\
Nanjing, China \\
\texttt{song-zhang@smail.nju.edu.cn} \\
\And
Amirhossein Habibian \\
Qualcomm AI Research \\
Amsterdam, Netherlands \\
\texttt{ahabibia@qti.qualcomm.com} \\
\And
Yawei Li\thanks{Corresponding author.} \\
Nanyang Technological University \\
Singapore \\
\texttt{yawei.li@ntu.edu.sg} \\
}

\iclrfinalcopy

\begin{document}

\maketitle
\lhead{Preprint}

\begin{abstract}

Smaller activation outliers do not necessarily imply better low-bit
quantization: their alignment with the quantizer matters.
We introduce \textbf{\pq{}}, a quantizer-aware rotation framework that
aligns the leading activation eigenspace with the constant group subspace
of asymmetric grouped INT4.
The affine offsets represent the energy in this subspace without widening
the range within the group.
We formulate rotation design as a Ky Fan trace maximization and derive a
closed-form solution that is \emph{provably optimal for this alignment
objective}.
Compact Householder transformations and their compact-WY representation
enable gradient-free construction and efficient application at both
foldable and online sites.
A predictive range law further connects unaligned activation energy and
group size to quantization-relevant variation.
Experiments on Llama, Qwen, and Mistral span dense models up to 70B
parameters and a 30B mixture-of-experts model.
Under W4A4KV4, \pq{} sets the state of the art on Llama-3.2-3B among the
compared methods in both perplexity and accuracy.
On Llama-3.1-70B, it attains \textbf{3.85} perplexity and
\textbf{72.46\%} average zero-shot accuracy, only 0.22 percentage points
below full precision.
In the deployment study on Llama-3.1-8B, our optimized implementation
achieves \textbf{$1.51\times$} prefill and \textbf{$1.22\times$} CUDA
Graph decode speedups over matched FP16 baselines, with
\textbf{56.34\%} lower decode peak memory and only \textbf{2.35\%}
additional Graph decode latency over Hadamard.
Code and models are available at \url{https://github.com/ForeverBlue816/PrismQuant} and \url{https://huggingface.co/collections/ForeverBlue/prsimquant}.

\end{abstract}

\section{Introduction}
\label{sec:intro}

4 bit weight-and-activation quantization offers a practical
route to reducing the memory and arithmetic costs of large language
model inference, but preserving accuracy remains challenging
\citep{ashkboos2024quarot,sun2024flatquant}.
A central obstacle is activation anisotropy: a few channels or
low-dimensional directions can dominate the quantization range,
leaving insufficient resolution for the remaining signal
\citep{dettmers2022gpt3,xiao2023smoothquant,sun2024massive}.
Rotations and other equivalent transforms mitigate this problem by
reshaping activation distributions while preserving the
full-precision function
\citep{lin2024duquant,hu2025ostquant,liu2025spinquant}.
Yet large magnitude alone does not determine quantization difficulty:
what matters is how the signal interacts with the quantizer's
representation. This motivates a complementary design question:
\emph{which activation directions does the quantizer already represent
efficiently, and how should a rotation exploit them?}

\begin{figure}[!t]
  \centering
  \includegraphics[width=\linewidth]{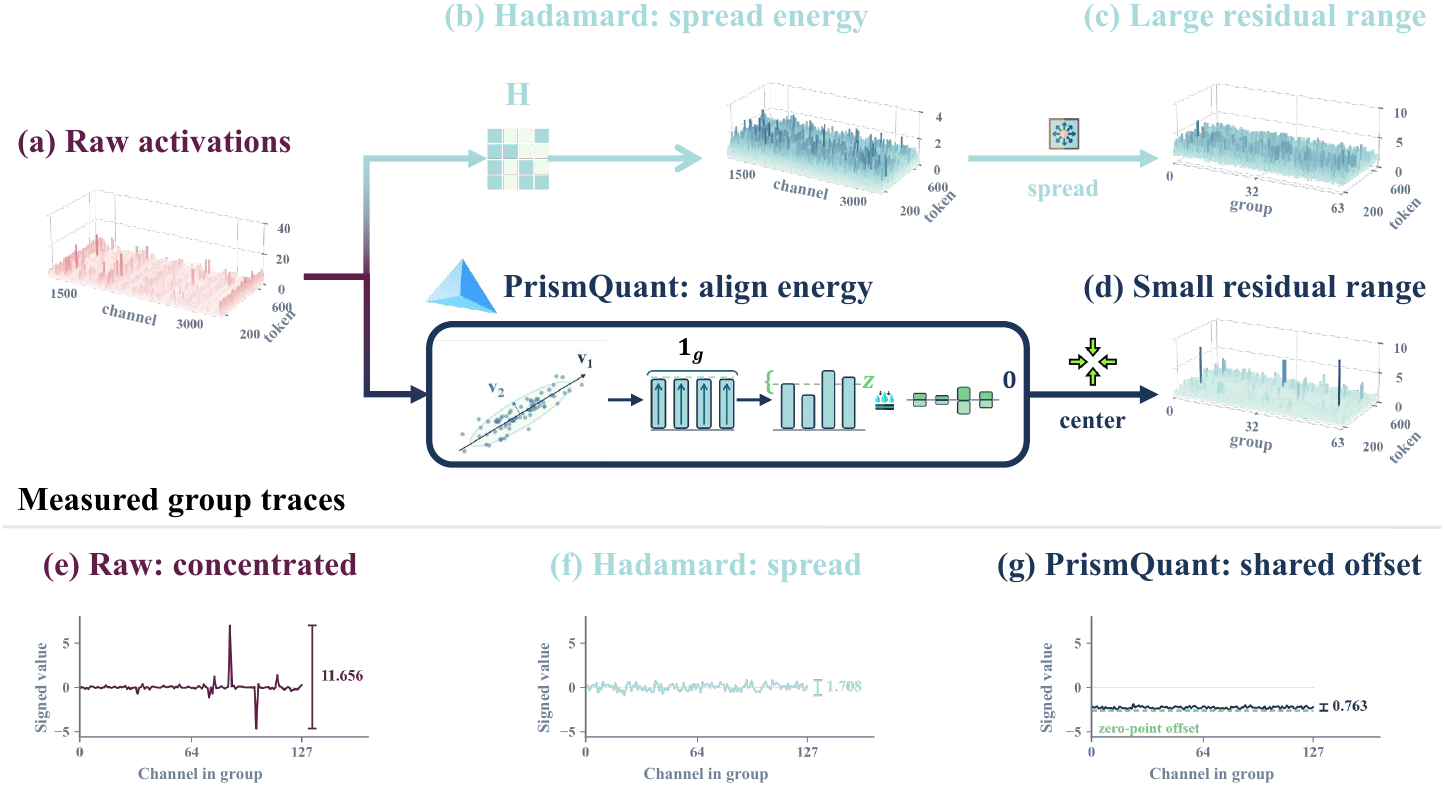}
  \caption{\textbf{Aligning activations with quantizer geometry.}
  Llama-3.2-3B layer 27 down-projection inputs, $g{=}128$.
  \textbf{(a--d)}~\pq{} reduces the mean within-group range from
  $1.76$ to $0.84$ relative to Hadamard under matched asymmetric
  INT4 quantization.
  \textbf{(e--g)}~Token/group traces illustrate the mechanism;
  the dashed line marks the stored affine offset.}
  \label{fig:teaser}
  \vspace{-0.5em}
\end{figure}

A grouped quantizer scales each group by its own extremes.
A component that is uniform within one group and absent from the others, therefore, raises no coordinate above that group's own scale, and in an asymmetric format its level is carried by the affine offset outright.
Across a $d$-dimensional activation with group size $g$, these
group-constant directions span a structured subspace of dimension $d/g$ that the quantizer tolerates best.
The quantizer's geometry, not only its precision, is thus something a transform can exploit.
We introduce \textbf{\pq{}}, a quantizer-aware rotation framework
that aligns dominant activation eigendirections with this
group-constant subspace through a closed-form rotation computed from
calibration statistics alone. As \Cref{fig:teaser} illustrates, this
alignment leaves substantially less within-group variation for
the same INT4 quantizer to resolve.

We formalize this principle in \Cref{sec:method}.
Let $\Sigma$ denote the activation second moment and $\mathcal{S}$
the group-constant subspace. Rotation design becomes the problem
of maximizing the expected energy projected onto $\mathcal{S}$
over orthogonal transforms. By the Ky Fan maximum principle
\citep{fan1949theorem,fan1950theorem}, the full-capacity optimum
maps the leading $d/g$ eigendirections of $\Sigma$ into
$\mathcal{S}$, yielding a solution that is
\emph{provably optimal for the alignment objective}.
In practice, we estimate the leading eigenspace from all calibration
tokens and realize the transform using Householder reflections in
compact $WY$ form \citep{schreiber1989storage}.
A rank parameter $k$ controls the alignment--cost trade-off,
without gradient-based training or quantization-aware distillation\citep{liu2025spinquant,chen2026gated}. The transform folds into adjacent
weights where possible; at non-foldable sites such as the
down-projection input, its compact representation avoids a dense
online rotation.

The same geometry reveals a second role for the size of the group. Smaller groups provide not only finer scale resolution but also more affine offsets, enlarging the subspace available for alignment. In \Cref{app:range_law}, we derive a two-factor range law that separates residual unaligned energy from the dependence of extreme-values on the size of the group
under a Gaussian residual approximation. This distinction matters: capturing more energy need not improve quantization if it requires coarser groups, since the captured energy depends only on the number of slots. Our metadata-matched ablations demonstrate this trade-off directly: among the tested configurations, allocating the same activation bit budget to finer groups is more effective than adding affine directions within larger groups. Alignment capacity and quantization granularity must therefore be considered together.

\begin{figure}[!t]
  \centering
  \includegraphics[width=\linewidth]{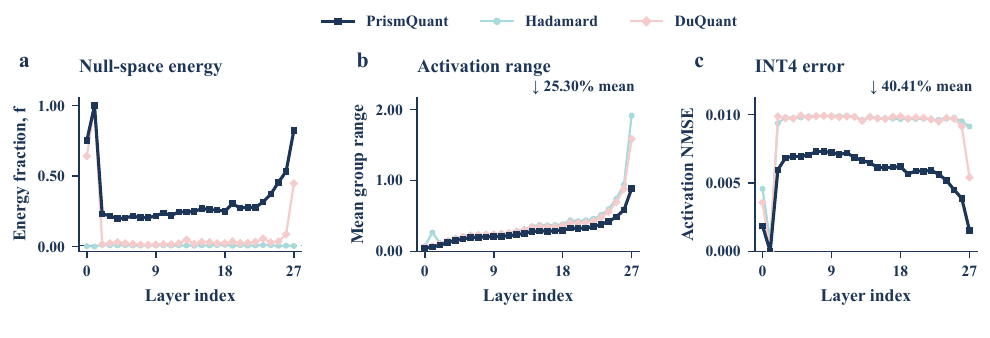}
  \caption{\textbf{Quantizer-aware alignment predicts INT4 error.}
  All 28 Llama-3.2-3B down-projection inputs, $g{=}128$,
  evaluated on identical tokens.
  \textbf{(a)}~Energy captured by the group-constant subspace.
  \textbf{(b,c)}~Within-group range and activation NMSE under
  matched asymmetric INT4; \pq{} reduces them by $25\%$ and
  $40\%$ on average relative to Hadamard.}
  \label{fig:perlayer}
  \vspace{-0.5em}
\end{figure}

Our experiments connect this geometry to local quantization error and
end-to-end model quality across the Llama, Qwen, and Mistral families,
from 0.6B to 70B parameters. Across all 28 down-projection inputs of Llama-3.2-3B, \pq{} reduces mean within-group range and activation NMSE by approximately $25\%$ and
$40\%$ relative to Hadamard under matched quantization settings (\Cref{fig:perlayer}); end to end, its strongest configuration reaches the lowest $8.58$ WikiText-2 perplexity and the highest eight-task zero-shot mean among compared methods (\textbf{61.23\%},\Cref{tab:llama_family}).
Controlled ablations in Section~\ref{subsec:ablations} probe the design
inward, at the alignment rank, the metadata budget, and the calibration
of sparsely routed experts; outward, at how the subspace is estimated;
and at the quantizer itself, where the gain survives a symmetric format
and moving the aligned level out of the offset's reach costs only a
quarter of it (Appendix~\ref{app:offset}).
Estimating the subspace from all calibration tokens rather than from a
few extreme ones recovers about a quarter more of the gain, and the gain
is insensitive to the calibration set, the eigensolver, and the signed
permutation (Appendix~\ref{app:additional_experiments}).
The same construction extends to mixtures of experts: with one rotation
per expert and the router untouched, Qwen3-30B-A3B recovers $85\%$ of
the accuracy that Hadamard loses (\Cref{tab:qwen3_moe}).
The rotation is also cheap to run: a two-kernel Tensor Core
implementation of the rank-$k$ correction, integrated into a packed-INT4
pipeline with CUDA Graph replay, adds $44$\,MB and $2.4\%$ decode latency
over Hadamard on Llama-3.1-8B while preserving the backend's $1.5\times$ prefill
throughput, $1.2\times$ decode speed, and $56\%$ lower peak memory than
FP16 (Appendix~\ref{app:prismquant_deployment}).

In summary, our contributions are:
\begin{itemize}[
    leftmargin=*,
    itemsep=0pt,
    parsep=0pt,
    topsep=1pt,
    partopsep=0pt
]
      \item \textbf{Quantizer-induced subspace alignment.}
      We formalize the group-constant geometry of grouped
     quantization and cast rotation design to maximize the activation energy
      captured by this subspace.
      Because the target subspace is fixed by the quantizer rather than
      estimated from outlier statistics, alignment becomes a single
      well-posed spectral problem with a closed-form optimum.
    
      \item \textbf{Spectral optimality with compact execution.}
      We characterize the alignment optimum through the Ky Fan principle
      and develop a training-free Householder realization with
      controllable rank, supporting both weight folding and online
      activation transforms.
      The online part reduces to a block Hadamard plus a rank-$k$
      correction, a structure that maps directly onto Tensor Core kernels
      and deploys in a packed-INT4 pipeline at negligible overhead.

    \item \textbf{Range law and end-to-end validation.}
      We prove that residual energy bounds the aggregate squared range,
      derive a two-factor law for the quantization step whose only
      approximation is a single measured crest-factor ratio, and test the
      mechanism with pre-registered ablations.
      In W4A4KV4 comparisons, \pq{} achieves the strongest
      results among the methods compared in various models and delivers
      real speedups and memory savings on commodity GPUs.
\end{itemize}

\section{Related Work}
\label{sec:related}

\paragraph{Activation outliers and smoothing.}
LLM activations are highly anisotropic, with a small number of channels,
tokens, or low-dimensional directions carrying extreme values
\citep{bondarenko2021understanding,dettmers2022gpt3,sun2024massive,wang2025demystifying}.
SmoothQuant \citep{xiao2023smoothquant} migrates the activation difficulty
into weights through per-channel scaling, OmniQuant
\citep{shao2024omniquant} jointly optimizes scaling and clipping,
Outlier Suppression+ \citep{wei2023outlier} adds a per-channel shift
that absorbs asymmetric outliers in a bias, and AWQ \citep{lin2024awq}
exploits activation statistics for weight-only quantization.
These channel-wise transformations rebalance ranges without mixing
information across channels.
Token-level isolation such as PrefixQuant \citep{chen2026prefixquant}
is complementary: with outlier tokens held in full precision, most of
\pq{}'s gain remains (Appendix~\ref{app:offset}).
\pq{} instead asks which directions the downstream quantizer efficiently represents
 and rotates dominant energy toward them.

\vspace{-0.8em}

\paragraph{Rotation-based quantization.}
QuaRot \citep{ashkboos2024quarot} uses fixed Hadamard transformations, SpinQuant \citep{liu2025spinquant} learns orthogonal rotations, and DuQuant \citep{lin2024duquant} combines channel reordering with structured block rotations. OSTQuant \citep{hu2025ostquant}, FlatQuant \citep{sun2024flatquant}, DFRot \citep{xiang2024dfrot}, KurTail \citep{akhondzadeh2025kurtail}, and DartQuant \citep{shao2026dartquant} further optimize scaling, rotation geometry, or activation distributions. These methods primarily optimize activation geometry.

A second line locates outlier directions from token statistics and treats them specially: ResQ \citep{saxena2024resq} keeps the high-variance directions in higher precision, and OffQ \citep{wang2026offq} selects the largest-norm token per sequence and rotates its principal direction into one channel per group, where the asymmetric zero-point absorbs it. Both lines take the activation as the object to be reshaped and the quantizer as given. \pq{} reverses the roles.
The asymmetric group quantizer already has a free subspace, spanned by the per-group constant directions; we treat that subspace as the target and ask which rotation carries the most activation energy into it.
The answer is a spectral problem with a closed-form optimum over the full second moment (Section~\ref{sec:method}), in which outliers enter only as energy the rotation captures rather than as directions to be found first. It is realized by a rank-$k$ compact-WY factor that folds into weights or runs online at the same cost (Appendix~\ref{app:prismquant_deployment}),
and the energy that remains unaligned is what our range law charges for,
as a function of group size and metadata budget (\Cref{app:range_law}, Figure~\ref{fig:geometry_range}).

\vspace{-0.8em}

\paragraph{Grouped asymmetric quantization.}
Group-wise scales and zero-points are widely used in low-bit weights, activations, and KV caches \citep{yuan2023rptq,lin2024awq,zhao2024atom,lin2025qserve}. KIVI \citep{liu2024kivi}, for example, shows that quantization granularity should reflect the statistical structure of keys and values. \pq{} highlights a complementary geometric consequence: in an asymmetric group, the affine offset induces a group-constant direction that can be deliberately targeted by an equivalent transform. Group size therefore controls not only metadata cost and local resolution, but also the dimension of the subspace available for alignment (Table~\ref{tab:app_ablations}).


\section{Method}
\label{sec:method}

A grouped asymmetric quantizer represents one direction in every group for
free: the constant vector spanned by its offset.
\pq{} rests on a single observation: an equivalent rotation can steer
the dominant activation energy into exactly these directions, so that what would otherwise set the quantization range is absorbed by metadata the format already pays for. Rather than flattening activations independently of the format, we therefore target the quantizer's own group-constant, range-neutral subspace. We derive the optimal spectral alignment, realize it through compact Householder transforms, and then describe Transformer integration and the metadata trade-off that follows. Proofs, numerical qualifications, and extended diagnostics are collected in \Cref{app:theoretical_foundations}.

\subsection{Quantizer-Induced Range-Null Subspace}
\label{sec:quantizer_geometry}

Let $X\in\mathbb{R}^{N\times d}$ hold token activations as rows. At a rotation
site every token is transformed by the same orthogonal $R$, so
$X'=XR^\top$, or $y=Rx$ for a single column-vector activation. Each transformed
token is split into $M=d/g$ contiguous groups of $g$ features, assuming $g$
divides $d$. Groups are per token: $d=8192$ and $g=128$ give 64 scales and
64 offsets for every token. Asymmetric INT4 encodes a nonconstant group
$y^{(j)}$ as
\begin{equation}
    q^{(j)}=\operatorname{clip}_{[0,15]}
    \left(\operatorname{round}\left(
        \frac{y^{(j)}-z_j\mathbf{1}_g}{s_j}
    \right)\right),
    \qquad
    \hat y^{(j)}=s_jq^{(j)}+z_j\mathbf{1}_g,
    \label{eq:asym_quant}
\end{equation}
with ideal parameters $z_j=\min y^{(j)}$ and
$s_j=\operatorname{range}(y^{(j)})/15$, where
$\operatorname{range}(v)=\max_i v_i-\min_i v_i$. The offset $z_j$ fixes the grid's origin and the scale $s_j$ its resolution;
both are stored in fp16, with a positive fallback scale for degenerate groups. The analysis below treats the metadata as exact
(\Cref{app:metadata_precision}). The range is blind to shared levels: if
$y^{(j)}=c_j\mathbf{1}_g+e^{(j)}$, its ideal scale depends only on
$\operatorname{range}(e^{(j)})$, while its offset becomes
$c_j+\min e^{(j)}$. Let $u_j=\mathbf{1}_{\mathcal I_j}/\sqrt g$ be the normalized
indicator of group $j$. These $M$ orthonormal vectors span the
\emph{group-constant subspace}
\begin{equation}
    \mathcal S=\operatorname{span}\{u_1,\ldots,u_M\},
    \qquad
    P_{\mathcal S}=UU^\top,
    \quad U=[u_1,\ldots,u_M],
    \label{eq:quantizer_subspace}
\end{equation}
whose projector replaces every group by its mean. Consequently,
\begin{equation}
    \operatorname{range}(y^{(j)})
    =\operatorname{range}\!\left(
        [(I-P_{\mathcal S})y]^{(j)}
    \right).
    \label{eq:range_residual_exact}
\end{equation}
We call $\mathcal S$ a \emph{range-null subspace}: its components do not contribute to the within-group range, rather than disappearing from the reconstructed
activation. Its $M=d/g$ degrees of freedom are represented through the
$16M$ bits of offsets already stored by the format. This decomposition makes explicit a property of the existing quantizer, it does not require explicit mean subtraction.

The shared level and the stored offset need not be identical: the latter
also contains the minimum of the remaining variation. Thus the existing
min--max quantizer already represents the aligned component without an
extra coefficient per token. A large activation magnitude is compatible
with a narrow group range when neighboring coordinates share that level.

Flattening alone does not ensure that dominant energy enters $\mathcal S$.
A Hadamard transform can reach group-constant directions, but it does not
select them according to the activation spectrum. For example, if the image
of an outlier channel is a signed pattern with zero mean in every group,
its projection onto $\mathcal S$ is zero and its full signed variation remains
for the INT4 grid. Other Hadamard columns can be group-constant, so zero
capture is not a universal property. For an isotropically oriented direction,
the expected energy fraction in $\mathcal S$ is $M/d=1/g$, providing the
reference level in \Cref{fig:perlayer}. \pq{} instead explicitly targets these
directions using calibration data.

\subsection{Optimal Alignment}
\label{sec:optimal_alignment}

We obtain dominant directions from the leading eigenvectors of the
uncentered second moment $\Sigma=\mathbb{E}[xx^\top]$, estimated offline
from $N_{\rm cal}$ calibration tokens as
$\widehat{\Sigma}=N_{\rm cal}^{-1}X_{\rm cal}^\top X_{\rm cal}$,
where $X_{\rm cal}\in\mathbb{R}^{N_{\rm cal}\times d}$ contains
calibration activations as rows.
We do not center: a persistent nonzero component also carries energy
that the quantizer must represent, and alignment can turn it into a
group-common component.
For the analysis, let $(\lambda_i,v_i)$ denote the eigenpairs of
$\Sigma$, ordered so that $\lambda_1\geq\cdots\geq\lambda_d$, and
write $V_k=[v_1,\ldots,v_k]$. Every token then decomposes as
\begin{equation}
    x=V_k a+r,
    \qquad
    a=V_k^\top x,
    \qquad
    r=(I-V_kV_k^\top)x,
    \label{eq:eigenspace_decomposition}
\end{equation}
where the directions are fixed at calibration and the coefficients $a$ vary
per token. Choose $k\leq M$ distinct targets
$U_k=[u_{j_1},\ldots,u_{j_k}]$ and require $RV_k=U_k$, absorbing fixed signs
into the eigenvector convention. Then
\begin{equation}
    Rx=
    \underbrace{U_k a}_{\text{group-common component}}
    +\underbrace{Rr}_{\text{remaining variation}}.
    \label{eq:aligned_decomposition}
\end{equation}
Under this alignment, the $i$-th leading component contributes a
token-dependent shared level $a_i/\sqrt g$ to group $j_i$.
The existing affine offset absorbs this shared level, leaving the
within-group range governed entirely by the residual $Rr$.
Any group-constant component of the residual likewise leaves the
range unchanged.
This structure is realized directly by rotating and quantizing $Rx$,
without explicitly computing or separately storing the coefficients $a$
at inference time.
\Cref{app:group_constant_geometry} illustrates this mechanism numerically.

\paragraph{Objective and its optimum.}
The energy delivered to the selected targets is
\begin{equation}
    \mathcal J_k(R)
    =\mathbb{E}\|U_k^\top Rx\|_2^2
    =\operatorname{Tr}\!\left(U_k^\top R\Sigma R^\top U_k\right),
    \qquad
    \max_{R^\top R=I}\ \mathcal J_k(R).
    \label{eq:alignment_objective}
\end{equation}
Because $R$ preserves total energy, maximizing $\mathcal J_k$ is equivalent
to minimizing the energy outside the selected targets,
$\operatorname{Tr}(\Sigma)-\mathcal J_k(R)$.

\begin{proposition}[Optimal spectral alignment]
\label{prop:optimal_alignment}
The maximum of \Cref{eq:alignment_objective} is
$\sum_{i=1}^k\lambda_i$. It is attained by any orthogonal $R$ mapping a leading
$k$-dimensional eigenspace of $\Sigma$ onto $\operatorname{span}(U_k)$.
\end{proposition}

The result follows from the Ky Fan maximum principle
\citep{fan1949theorem,fan1950theorem}: $\operatorname{Tr}(Q^\top\Sigma Q)$
over $d\times k$ matrices with orthonormal columns is maximized by a leading
eigenspace, and $Q=R^\top U_k$ ranges over exactly that set. The selected
alignment is therefore provably optimal for this objective. At $k=M$, it
minimizes total energy outside $\mathcal S$; for $k<M$, the guarantee applies
to the selected $k$-dimensional target. 

The optimal subspace, rather than a unique orthogonal matrix, is the
object of this construction. Eigenvalue ties may admit several equally
valid leading eigenspaces. In practice, alignment of an orthonormal
estimate $\widehat V_k$ captures
$\operatorname{Tr}(\widehat V_k^\top\Sigma\widehat V_k)$ in the selected
targets. This separates the quality of the calibrated directions from
their structured realization: estimating the empirical moment and
constructing its rotation are distinct steps, and neither requires
optimizing a task loss.

\begin{figure}[!hb]
    \centering
    \captionsetup{font=small,skip=4pt}
    \includegraphics[width=\linewidth]{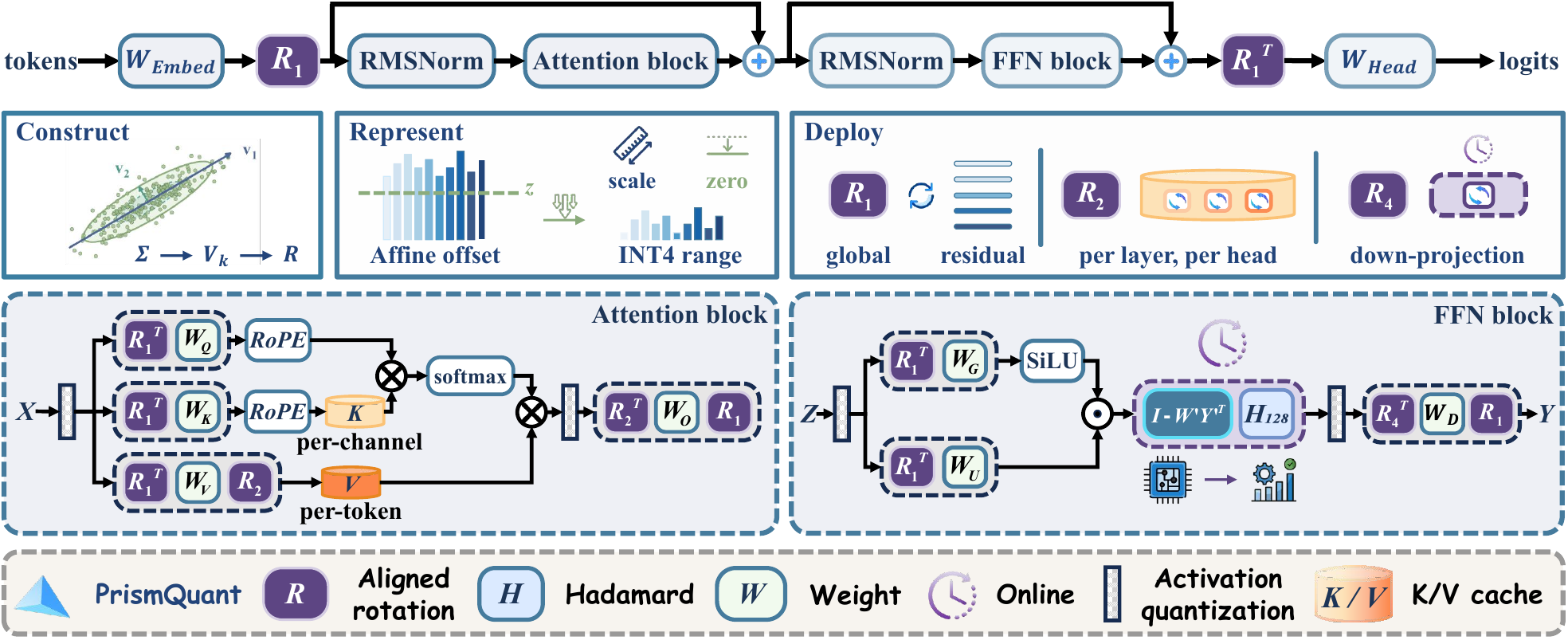}
    \caption{\textbf{Overview of \pq{}.}
    Calibration takes the leading eigenvectors $V_k$ of the second moment
    $\Sigma$ and builds a rotation that turns them into per-group shared
    levels, which the affine offset $z$ represents outside the INT4 range. $R_1$ is folded into the embedding, head, and all residual-facing weights; $R_2$ is applied after $W_V$ and its transpose folded into $W_O$ so the V cache is quantized in the aligned basis; at the down-projection input only the rank-$k$ correction $I-W'Y'^\top$ and the block Hadamard $H_{128}$ run online (clock), with $R_4^\top$ folded into $W_D$.
    The schematic uses right-acting row-vector rotations.}
    \label{fig:method}
\end{figure}

\subsection{Structured Construction}
\label{sec:structured_rotation}

We realize the alignment without storing a dense transform:
\begin{equation}
    R=H_gD\Pi G,
    \qquad
    G=I-WY^\top,
    \quad W,Y\in\mathbb{R}^{d\times k}.
    \label{eq:prismquant_factorization}
\end{equation}
$G$ maps leading eigenvectors to coordinate anchors,
$\Pi$ arranges coordinates across groups to balance
residual energy while keeping these anchors fixed, $D$ applies fixed signs, and $H_g$ converts each anchor into a group constant direction. Here $H_g$ denotes the block-diagonal transform with a normalized size-$g$ Walsh--Hadamard matrix in each block\citep{fino1976unified,ashrafi2017walsh}.

\vspace{-0.5em}

\paragraph{Alignment by Householder reflections.}
The construction first maps spectral
directions to coordinate anchors. Set $G_0=I$ and
$t_i=1+(i-1)g$ for $i=1,\ldots,k$. At step $i$,
$b_i=G_{i-1}v_i$ is orthogonal to the previously fixed anchors.
For $\delta_i=b_i-e_{t_i}\neq0$, define
\begin{equation}
    h_i=\frac{\delta_i}{\|\delta_i\|_2},
    \qquad \mathcal H_i=I-2h_i h_i^\top,
    \qquad G_i=\mathcal H_iG_{i-1}.
    \label{eq:householder_step}
\end{equation}
The reflection sends $b_i$ to $e_{t_i}$ while preserving earlier
anchors; after at most $k$ reflections, $G=G_k$ aligns every selected
direction. The numerical skip rule and induction proof are given in
\Cref{app:householder_construction}.

The compact $WY$ representation \citep{schreiber1989storage} collects the
reflectors into two thin factors. Using the column-vector convention in
\Cref{eq:prismquant_factorization}, batched application is
\begin{equation}
    XG^\top=X-(XY)W^\top,
    \qquad XY\in\mathbb{R}^{N\times k}.
    \label{eq:wy_apply}
\end{equation}
The activation dimension remains $d$: $G$ is orthogonal and full rank,
preserving all information and the Euclidean norm of each activation.
Only the correction $I-G=WY^\top$ has rank at most $k$.
Applying this correction through the two thin factors requires
$\mathcal O(dk)$ operations per token and $\mathcal O(dk)$ storage
for $W$ and $Y$, compared with $\mathcal O(d^2)$ computation and
storage for a dense transform.

\vspace{-0.5em}

\paragraph{Anchor placement and residual balancing.} The $k$ aligned anchors occupy the first coordinates of the first $k$ groups. When $k<M$, the first coordinates of the remaining groups are filled by the lowest-energy available coordinates. We estimate coordinate energies after $G$ from calibration activations, corresponding to the diagonal of $G\Sigma G^\top$. All other coordinates are sorted by decreasing energy and assigned greedily to the group with the lowest accumulated residual energy, subject to its $g-1$ non-anchor slots; ties are resolved deterministically. The low-energy fillers keep reduced-rank ablations from unintentionally assigning additional high-energy coordinates to unused constant slots. These choices preserve the selected eigenspace alignment.

\vspace{-0.5em}

\paragraph{Block mixing and calibration.}
Within each group, a normalized Walsh--Hadamard matrix maps $e_1$ to $\mathbf{1}_g/\sqrt g$, while its remaining columns are orthogonal to that constant direction.
 Therefore, the complete construction satisfies $Rv_i=\pm u_i$ for the
selected directions, converting the aligned anchors into shared levels
and mixing the residual within each group.
This final stage costs $\mathcal O(d\log g)$ per token, bringing the
total transform cost to $\mathcal O(dk+d\log g)$.
To obtain the spectral directions used in this construction, we employ
a randomized eigensolver \citep{halko2011finding} at wide activation
sites, avoiding explicit formation of the full second moment.
Our implementation uses three passes with orthogonalization and
Rayleigh--Ritz extraction, while the smaller value moments within
each attention head are formed explicitly and diagonalized directly.
The coordinate energies for the permutation are estimated in a separate
calibration step, after which all transform factors and seeded signs
are fixed for evaluation without gradient-based training.
We record Ritz and anchor residuals separately to distinguish
eigenspace approximation from alignment accuracy.
\Cref{app:method_calibration} details the pass accounting,
the sketch parameters and the numerical checks.

\subsection{Deployment in the Transformer}
\label{sec:deployment}

For a consuming linear map $A$, $Ax=(AR^\top)(Rx)$, and $AR^\top$ is
precomputed offline. Whether the forward operation $Rx$ disappears depends on the upstream computation, as distinguished in \Cref{fig:method}.

\emph{Residual-stream inputs ($R_1$).}
A single rotation $R_1$ re-expresses the residual stream in one shared basis for the whole network. Because every layer reads from and writes to the same stream, the change of basis is absorbed entirely into the weights: once the RMSNorm gains are folded into the adjacent reading weights, each residual-reading matrix becomes $A_{\rm read}R_1^\top$, each residual-writing matrix $R_1A_{\rm write}$, and the embedding and output head transform accordingly, so this site needs no online rotation. Since $R_1$ is shared, we calibrate it on a second moment pooled over all layers from the attention-input RMSNorm outputs, weighting every token equally; the resulting rotation is optimal for this pooled objective, even though individual layers would each prefer a different one. Per-layer diagnostics report how much alignment each site retains under the shared rotation.

\emph{Down-projection input ($R_4$).}
For $h=\operatorname{SiLU}(A_{\rm gate}x)\odot(A_{\rm up}x)$, a rotation
cannot be moved through the element-wise gate, so $R_4$ is calibrated on each
layer's down-projection input, applied online after the gate, and its
transpose folded into $A_{\rm down}R_4^\top$.
Its signed permutation still can be folded: with $T=D\Pi$, setting
$A_{\rm gate}\leftarrow\Pi A_{\rm gate}$ and
$A_{\rm up}\leftarrow D\Pi A_{\rm up}$ yields $h_T=Th$ directly, where only the up branch receives signs because SiLU does not commute with sign flips.
What remains online is:
\begin{equation}
    R_4h=H_g\widetilde G h_T,
    \qquad
    \widetilde G=TGT^\top,
    \label{eq:folded_online_rotation}
\end{equation}
a block Hadamard and a rank-$k$ correction, which a two-kernel Tensor Core implementation executes without any online gather; see
\Cref{app:prismquant_deployment,fig:app_prism_efficiency} for details.

\emph{Values and the KV cache ($R_2$).}
Values receive a calibrated head-dimensional rotation $R_2$ after the value
projection, with inverse compensation folded into the corresponding
output-projection head blocks, including GQA expansion. At each layer,
the value second moment is pooled across tokens and KV heads, and one
rotation is shared across those heads. For the 128-dimensional heads used
here, a single head is one quantization group, so $R_2$ aligns one leading
direction ($k_{R_2}=1$). The reference evaluation applies this forward
rotation through a value-projection hook.

Keys receive no additional quantization rotation after RoPE. They are
quantized per channel across completed groups of 32 tokens; values are
quantized per token across each head's feature dimension. The KIVI-style
policy retains the current residual key chunk (up to 32 tokens) and the
most recent 32 value tokens at full precision~\citep{liu2024kivi}. Thus, keys and values use different
grouping axes, while value alignment
exploits the per-head affine offset.
The nominal rank $k$ controls $R_1$ and $R_4$, capped by each site's number
of group-constant directions; $R_2$ uses its head-specific rank above.

\subsection{Range and Metadata}
\label{subsec:range_metadata}

The objective controls residual energy, whereas the quantizer responds to
range.
The two are linked by the deterministic inequality
$\sum_{j}\operatorname{range}(y^{(j)})^2\le 2\|(I-P_{\mathcal S})y\|_2^2$,
which bounds the aggregate squared range but does not fix the mean step.
Assuming the mixed residual keeps its normalized shape as energy is
removed, its range scales with the amplitude $\sqrt{1-f_k}$, with
$f_k=\sum_{i\le k}\lambda_i/\operatorname{Tr}(\Sigma)$ the aligned
energy fraction, which motivates the baseline-relative predictor
\begin{equation}
    \widehat s(g,k)=s_{\rm H}(g)\sqrt{1-f_k},
    \label{eq:range_predictor_main}
\end{equation}
where $s_{\rm H}(g)$ is the measured mean step of the paired Hadamard
reference and no coefficient is fitted.
It is an approximation, not a bound: an exact identity
(\Cref{app:range_prediction}) isolates its error in one shape factor,
the ratio of residual crest factors after and before alignment, measured
near one at the q/k/v input and $0.87$--$0.92$ at the down-projection
input.

The same $g$ governs a second resource, \textbf{the metadata budget}.
With one fp16 scale and offset per group, codes and metadata occupy
$4+32/g$ bits per value; halving $g$ doubles both the number of local
scales and the dimension of $\mathcal S$, so finer groups buy resolution
and alignment capacity together at a metadata cost, whereas raising $k$
leaves the format budget untouched and costs factor storage
\textbf{and online work} instead.
Extended-affine controls, which add $m-1$ fixed within-group directions at $4+16(m+1)/g$ bits, separate the two: at a shared $4.25$-bit budget, one direction per group of 128 beats three per group of 256
\textbf{on both Llama models} (\Cref{tab:app_ablations}a).
\Cref{app:method_range} derives the law, gives the metadata accounting,
and reports the range diagnostics (\Cref{fig:geometry_range}).

\providecommand{\pqtdelta}[1]{{\scriptsize\textcolor{PQTGreen}{($#1$)}}}
\providecommand{\pqtdefault}{\textcolor{PQTGreen}{\scriptsize default}}
\providecommand{\pqtworse}[1]{\textcolor{PQTWarm}{$#1$}}
\providecommand{\pqtzero}[1]{\textcolor{PQTGreen}{$#1$}}

\providecolor{PQTOursDeep}{HTML}{D3E1EC}
\providecommand{\pqtdelta}[1]{{\scriptsize\textcolor{PQTGreen}{($#1$)}}}
\providecommand{\pqtdefault}{\textcolor{PQTGreen}{\scriptsize default}}
\providecommand{\pqtworse}[1]{\textcolor{PQTWarm}{$#1$}}
\providecommand{\pqtzero}[1]{\textcolor{PQTGreen}{$#1$}}

\section{Experiments}
\label{sec:experiments}

We evaluate W4A4KV4 post-training quantization on Llama-3.2-3B,
Llama-3.1-8B, and Llama-3.1-70B, the mixture-of-experts
Qwen3-30B-A3B-Base, and, in the appendix, the Qwen3 dense family from
0.6B to 8B and Mistral-7B-v0.3
\citep{grattafiori2024llama,yang2025qwen3}.
Activations use the asymmetric group quantizer of
Section~\ref{sec:method} with $g=128$ ($4.25$ bits per value);
weights use GPTQ INT4 and the KV cache follows KIVI
\citep{liu2024kivi}.
Rotations are applied at the standard sites: $R_1$ on the residual
stream, $R_2$ after the value projection with its transpose folded into
the output projection, and $R_4$ online before the down projection;
GPTQ runs with the rotations in place.
The Hadamard baseline places a dense Hadamard at every site; \pq{}
keeps everything else fixed and replaces the Hadamards with the
null-space-aligned rotations of Section~\ref{sec:method}, with $k$
aligned directions per site and $k=\max$ using every group slot.
Rotation statistics and GPTQ share 128 calibration sequences of
2048 tokens. All reported \pq{} results are averaged over three random seeds. We report WikiText-2 and C4 perplexity, zero-shot accuracy on eight
tasks, and, for the Qwen3 family, MMLU, MMLU-Redux, and GSM8K.
Algorithm~\ref{alg:prismquant} (Appendix~\ref{app:algorithm}) lists
the full procedure and Appendix~\ref{app:setup} the complete
protocol.

\subsection{Dense Models}
\label{subsec:dense}


\begin{table}[t]
\centering
\captionsetup{font=small,skip=4pt}

\caption{\textbf{W4A4KV4 quantization across the Llama family.}
WikiText-2 perplexity ($\downarrow$) and zero-shot accuracy
(\%, $\uparrow$); Avg. averages the eight displayed tasks.
Published rows and the 3B/8B bf16 references follow
\citet{wang2026offq} (3B) and \citet{he2025base} (8B/70B); the 70B
reference, Hadamard, and \pq{} rows are measured in our pipeline.
Bold marks the best quantized result per column within each model;
\pqtours{} rows are shaded light for $k=8$ and darker for $k=\max$.}

\label{tab:llama_family}
\label{tab:llama70b}

\fontsize{8pt}{9pt}\selectfont
\setlength{\tabcolsep}{1.8pt}
\renewcommand{\arraystretch}{1.00}
\setlength{\extrarowheight}{0.3pt}

\setlength{\aboverulesep}{1.8pt}
\setlength{\belowrulesep}{1.8pt}
\arrayrulecolor{black}

\newcolumntype{Q}{>{\raggedleft\arraybackslash}X}

\begin{tabularx}{\linewidth}{lQQQQQQQQQQ}
\toprule

& &
\multicolumn{8}{c}{\textbf{Zero-shot accuracy} $\uparrow$}
& \\

\cmidrule(lr){3-10}

\textbf{Method}
& \mbox{\textbf{PPL}\,$\downarrow$}
& ARC-c
& ARC-e
& BoolQ
& HellaS
& OBQA
& PIQA
& SIQA
& WinoG
& \mbox{\textbf{Avg.}\,$\uparrow$} \\

\midrule


\rowcolor{PQTRef}
\textcolor{PQTInk}{\textbf{Llama-3.2-3B} (bf16)}
& 7.80
& 46.20 & 71.70 & 73.10 & 73.70
& 43.40 & 77.40 & 47.20 & 69.10
& 62.73 \\

QuaRot\pqtvenue{NeurIPS}{24}
& 10.10
& 38.60 & 59.00 & 65.90 & 66.50
& 35.80 & 74.40 & 43.10 & 65.20
& 56.06 \\

SpinQuant\pqtvenue{ICLR}{25}
& 9.20
& 38.90 & 64.80 & 68.00 & 69.10
& 39.40 & 74.90 & 45.10 & 62.90
& 57.89 \\

KurTail\pqtvenue{EMNLP-F}{25}
& 9.00
& 42.20 & 66.70 & 69.80 & 68.80
& 39.80 & \pqtbest{75.60} & 44.80 & 64.60
& 59.04 \\

ResQ\pqtvenue{ICML}{25}
& 8.80
& 43.10 & 65.60 & 68.80 & 70.50
& 38.40 & 75.10 & 45.60 & 64.80
& 58.99 \\

\rowcolor{PQTOffQ}
OffQ\pqtvenue{arXiv}{26}
& 8.78
& \pqtbest{44.80}
& \pqtbest{70.41}
& 71.62
& 71.49
& 39.60
& 75.52
& 45.65
& 67.32
& 60.80 \\

\pqtdashline

Hadamard
& 9.04
& 43.00 & 67.63 & 66.97 & 71.18
& 40.00 & 74.27 & 45.60 & 65.67
& 59.29 \\

\rowcolor{PQTOurs}
\pqtours{} ($k=8$)
& 8.60
& 43.43
& 69.91
& 74.07
& 71.59
& 40.20
& 74.97
& 45.50
& 66.77
& 60.80 \\

\rowcolor{PQTOursDeep}
\pqtours{} ($k=\max$)
& \pqtbest{8.58}
& 43.69
& 70.24
& \pqtbest{74.16}
& \pqtbest{71.60}
& \pqtbest{41.00}
& 75.08
& \pqtbest{45.91}
& \pqtbest{68.19}
& \pqtbest{61.23} \\

\midrule


\rowcolor{PQTRef}
\textcolor{PQTInk}{\textbf{Llama-3.1-8B} (bf16)}
& 6.23
& 53.50 & 81.19 & 82.08 & 78.90
& 44.80 & 81.23 & 47.19 & 73.56
& 67.81 \\

QuaRot\pqtvenue{NeurIPS}{24}
& 7.82
& 46.08 & 72.31 & 77.34 & 74.46
& 42.40 & 76.82 & 44.37 & 68.51
& 62.79 \\

SpinQuant\pqtvenue{ICLR}{25}
& 7.51
& 47.87 & 76.05 & 76.76 & 74.63
& 43.20 & 79.27 & 45.70 & 67.17
& 63.83 \\

OSTQuant\pqtvenue{ICLR}{25}
& 7.40
& 47.44 & 75.34 & 78.84 & 75.48
& 41.80 & 78.24 & 47.13 & 68.51
& 64.10 \\

\rowcolor{PQTBaseQ}
BASE-Q\pqtvenue{arXiv}{25}
& 7.17
& 49.15 & 77.19 & 78.56 & 76.41
& 42.40 & 79.16 & 45.85 & 69.22
& 64.74 \\

\pqtdashline

Hadamard
& 7.21
& 50.34
& 78.24
& 80.03
& 76.13
& 42.60
& \pqtbest{79.33}
& 46.72
& 69.53
& 65.37 \\

\rowcolor{PQTOurs}
\pqtours{} ($k=8$)
& 6.99
& 48.89
& \pqtbest{79.00}
& 80.12
& \pqtbest{76.95}
& \pqtbest{44.20}
& 78.13
& \pqtbest{47.24}
& 68.35
& 65.36 \\

\rowcolor{PQTOursDeep}
\pqtours{} ($k=\max$)
& \pqtbest{6.92}
& \pqtbest{50.77}
& 77.86
& \pqtbest{80.55}
& 76.72
& 43.00
& 78.84
& 45.50
& \pqtbest{70.24}
& \pqtbest{65.43} \\

\midrule


\rowcolor{PQTRef}
\textcolor{PQTInk}{\textbf{Llama-3.1-70B} (bf16)}
& 2.81
& 61.09 & 81.69 & 87.16 & 85.66
& 47.80 & 84.22 & 51.79 & 82.00
& 72.68 \\

QuaRot\pqtvenue{NeurIPS}{24}
& 5.31
& 57.51 & 80.35 & 83.27 & 81.45
& 44.80 & 81.66 & 45.60 & 75.77
& 68.80 \\

SpinQuant\pqtvenue{ICLR}{25}
& 4.74
& 59.13 & 82.28 & 84.37 & 82.24
& 46.00 & 82.21 & 47.24 & 76.40
& 69.98 \\

\rowcolor{PQTBaseQ}
BASE-Q\pqtvenue{arXiv}{25}
& 4.17
& 60.49
& \pqtbest{84.13}
& 83.12
& 83.10
& 47.40
& 83.24
& 47.85
& 77.43
& 70.85 \\

\pqtdashline

Hadamard
& 4.22
& 60.90
& 83.30
& 85.00
& 83.80
& 47.20
& 83.20
& 48.60
& 78.90
& 71.36 \\

\rowcolor{PQTOurs}
\pqtours{} ($k=8$)
& \pqtbest{3.85}
& \pqtbest{61.86}
& 83.59
& \pqtbest{86.73}
& \pqtbest{84.64}
& \pqtbest{47.60}
& \pqtbest{83.79}
& \pqtbest{51.02}
& \pqtbest{80.43}
& \pqtbest{72.46} \\

\rowcolor{PQTOursDeep}
\pqtours{} ($k=\max$)
& 3.89
& 61.52
& 83.63
& 86.42
& 84.59
& 46.60
& 83.57
& 49.08
& 80.35
& 71.97 \\

\bottomrule
\end{tabularx}
\end{table}

\pq{} consistently achieves the lowest perplexity among all displayed
quantized methods across the three Llama scales
(Table~\ref{tab:llama_family}).
Against the metadata-matched Hadamard baseline, it closes
$\mathbf{37\%}$, $\mathbf{30\%}$, and $\mathbf{26\%}$ of the
perplexity gap to bf16 on 3B, 8B, and 70B, respectively.
It also recovers most of the average accuracy loss on 3B and 70B,
gaining $\mathbf{1.94}$ points ($59.29\to61.23$) and
$\mathbf{1.10}$ points ($71.36\to72.46$).
On 70B, it improves all eight tasks over Hadamard and comes within
$\mathbf{0.22}$ points of bf16.
Compared with published baselines, \pq{} surpasses OffQ on 3B by
$\mathbf{0.43}$ accuracy points with $0.20$ lower PPL, and BASE-Q
by $0.69$ points on 8B and $\mathbf{1.61}$ points with $0.32$ lower
PPL on 70B.
The preferred rank varies with model scale: $k=\max$ yields the
best perplexity and average accuracy on 3B and 8B, while $k=8$
leads on 70B.

\subsection{Mixture of Experts}
\label{subsec:moe}


\begin{table}[t]
\centering
\captionsetup{font=small,skip=3pt}

\caption{\textbf{W4A4KV4 quantization of Qwen3-30B-A3B-Base.}
WikiText-2 and C4 perplexity ($\downarrow$) and zero-shot accuracy
(\%, $\uparrow$) on the eight-task suite of Table~\ref{tab:llama_family}.
All rows are evaluated in our pipeline with three seeds; the router is
kept in bf16, and every expert receives its own $R_4$ ($k=6$, its
full slot count), so the two \pqtours{} rows differ only in the rank
of $R_1$.
Bold marks the best quantized result per column.}

\label{tab:qwen3_moe}

\begingroup
\fontsize{7pt}{8pt}\selectfont
\setlength{\tabcolsep}{1.7pt}
\renewcommand{\arraystretch}{1}
\setlength{\aboverulesep}{1.8pt}
\setlength{\belowrulesep}{1.8pt}
\arrayrulecolor{black}

\begin{tabularx}{\linewidth}{
    l*{11}{>{\raggedleft\arraybackslash}X}
}
\toprule

& \multicolumn{2}{c}{\textbf{PPL} $\downarrow$}
& \multicolumn{8}{c}{\textbf{Zero-shot accuracy} $\uparrow$}
& \\[-1pt]

\cmidrule(lr){2-3}
\cmidrule(lr){4-11}

\textbf{Method}
& WT2 & C4
& ARC-c & ARC-e & BoolQ & HellaS & OBQA & PIQA & SIQA & WinoG
& \mbox{\textbf{Avg.}\,$\uparrow$} \\

\midrule

\rowcolor{PQTRef}
\textcolor{PQTInk}{\textbf{Qwen3-30B-A3B} (bf16)}
& 6.11 & 10.70
& 57.68 & 79.55 & 81.41 & 81.40 & 44.00 & 80.96 & 52.15 & 72.53
& 68.71 \\

Hadamard
& 6.68 & 11.47
& 55.38 & 79.04 & 81.41 & 79.24 & 44.80 & 79.33 & 50.26 & 70.32
& 67.47 \\

\pqtdashline

\rowcolor{PQTOurs}
\pqtours{} ($k=8$)
& 6.45 & 11.19
& \pqtbest{57.25} & 79.29 & 81.90 & \pqtbest{80.08} & \pqtbest{45.40}
& \pqtbest{80.30} & 51.28 & \pqtbest{72.61}
& \pqtbest{68.52} \\

\rowcolor{PQTOursDeep}
\pqtours{} ($k=\max$)
& \pqtbest{6.42} & \pqtbest{11.18}
& 55.63 & \pqtbest{79.84} & \pqtbest{83.55} & 79.60 & 44.40
& 79.60 & \pqtbest{51.94} & 71.27
& 68.23 \\

\bottomrule
\end{tabularx}
\endgroup
\end{table}

The advantage transfers to sparsely routed models
(Table~\ref{tab:qwen3_moe}).
Qwen3-30B-A3B routes each token to 8 of 128 experts per layer; we
quantize every expert with its own $R_4$, keep the router in bf16 with
$R_1$ folded into its input so that routing is unchanged, and calibrate
cold experts with the shrinkage estimator of
Eq.~\ref{eq:expert_covariance_shrinkage} (Appendix~\ref{app:setup}).
Hadamard costs $0.57$ WikiText-2 PPL, $0.77$ C4 PPL, and $1.24$
accuracy points against bf16; \pq{} recovers $\mathbf{46\%}$ and
$\mathbf{38\%}$ of the two perplexity gaps at $k=\max$ and
$\mathbf{1.05}$ of the $1.24$ points at $k=8$, ahead of Hadamard on all
eight tasks and only $\mathbf{0.19}$ below bf16, a gain of $2.4$
standard errors.
The two \pq{} rows share their expert rotations and differ only in the
rank of $R_1$; their $0.29$-point difference is within one standard
error.

\subsection{Ablation Studies}
\label{subsec:ablations}

\begin{table}[t]
\centering
\captionsetup{font=small,skip=4pt}
\caption{\textbf{Alignment rank.}
Activation-only recovery ($\%$, $\uparrow$) of the Hadamard-to-bf16
WikiText-2 perplexity gap by \pq{} at rank $k$; the first row lists the
gap itself.
Weights and KV states remain in bf16; ranks are capped per site by the
number of group slots.
Bold marks the largest recovery; row shading darkens with $k$.}
\label{tab:rank_ablation}
\begingroup
\fontsize{7pt}{8pt}\selectfont
\setlength{\tabcolsep}{1.8pt}
\renewcommand{\arraystretch}{1.00}
\setlength{\extrarowheight}{0.3pt}
\setlength{\aboverulesep}{1.8pt}
\setlength{\belowrulesep}{1.8pt}
\arrayrulecolor{black}
\begin{tabularx}{\linewidth}{
    >{\raggedright\arraybackslash}p{0.40\linewidth}
    *{3}{>{\raggedleft\arraybackslash}X}
}
\toprule
\rowcolor{PQTBand}
\textbf{Configuration}
& \shortstack[r]{Llama\mbox{-}3.2\mbox{-}3B}
& \shortstack[r]{Llama\mbox{-}3.1\mbox{-}8B}
& \shortstack[r]{Qwen3\mbox{-}8B} \\
\midrule
Hadamard gap, $P_{\mathrm{Had}}-P_{\mathrm{bf16}}$ (PPL)
& \pqtworse{\mathbf{+0.154}}
& \pqtworse{\mathbf{+0.142}}
& \pqtworse{\mathbf{+0.442}} \\
\pqtdashline
\rowcolor{PQTOurs}
\pqtours{} ($k=8$)
& 38.59 & 35.53 & 59.12 \\
\rowcolor{PQTOurs!75!PQTOursDeep}
\pqtours{} ($k=16$)
& 38.30 & 35.74 & 51.97 \\
\rowcolor{PQTOurs!50!PQTOursDeep}
\pqtours{} ($k=32$)
& 41.88 & 40.23 & 85.57 \\
\rowcolor{PQTOurs!25!PQTOursDeep}
\pqtours{} ($k=64$)
& 40.83 & 41.84 & 99.85 \\
\rowcolor{PQTOursDeep}
\pqtours{} ($k=\max$)
& \pqtbest{42.67} & \pqtbest{43.12} & \pqtbest{99.95} \\
\bottomrule
\end{tabularx}
\endgroup
\end{table}

\paragraph{Alignment rank.}
A small rank captures most of the gain on Llama, and the optimum is
model dependent.
Table~\ref{tab:rank_ablation} reports the fraction of the
Hadamard-to-bf16 perplexity gap that \pq{} closes when only the q/k/v
and down-projection inputs are quantized.
On Llama-3.2-3B and Llama-3.1-8B, $k=8$ recovers $\mathbf{38.59\%}$ and
$\mathbf{35.53\%}$ of the gap, and $k=\max$ only $42.67\%$ and $43.12\%$.
On Qwen3-8B, whose Hadamard gap is three times larger and whose
down-projection inputs are dominated by group-constant directions
(Appendix~\ref{app:local_visualizations}), recovery rises from
$59.12\%$ to $\mathbf{99.95\%}$: alignment removes the whole
activation-quantization penalty.
Recovery is not monotone at intermediate ranks: the alignment objective
is solved exactly at every $k$, but the perplexity it induces is not.
We therefore use $k=8$ as the operating point and report $k=\max$ as the
alignment optimum.
\vspace{-0.5em}

\vspace{-0.5em}
\paragraph{Further ablations.}
Three more ablations, reported in
Appendix~\ref{app:additional_experiments}, probe the remaining design
choices: how a fixed activation bit budget should be split between group
size and represented directions, whether the moe result
depends on how cold experts are calibrated, and which calibration tokens
should define the subspace, including the outlier-token selection used by prior work.

\vspace{-0.5em}

\section{Conclusion}
\label{sec:conclusion}

\vspace{-0.5em}
\pq{} starts from the structure of the quantizer: the directions a
grouped format tolerates best are fixed before any data is seen, and
rotation design reduces to a spectral problem with a closed-form
optimum, a rank-$k$ Householder realization, and a range law that ties
the residual step to unaligned energy and group size.
The construction is correspondingly simple, insensitive to how the
residual is permuted, how the calibration set is chosen, or how
precisely the eigenspace is solved, and it extends without change from
dense models to sparsely routed experts.
Beyond this work, an INT4 GEMM that consumes group-wise asymmetric
activations natively would turn the transform's measured overhead into
a checkpoint-level system, and the same alignment applies to any format
that scales by group extrema, including block floating-point and
microscaling variants.
\clearpage

\clearpage

\subsection*{AI use statement}

In this work, we used generative AI tools for manuscript drafting and figure preparation. We have not used generative AI tools for generating novel research ideas, writing experimental code, or generating data samples for calibration and benchmarking, and other tasks with required disclosure are not applicable to this work. Additionally, we used generative AI tools for language editing and polishing. 

We have reviewed all AI-assisted work. Specifically, the authors manually verified all drafted text for scientific accuracy, reviewed all generated figures to ensure they correctly represent our methodology, and checked all citations and references to prevent hallucinations. We take full responsibility for the final content of this work, including text, claims, results, and artifacts produced with the aid of generative AI.

\subsection*{Ethics statement}

This work studies the quantization of existing language models using publicly
available checkpoints and benchmark datasets, without conducting new
human-subject studies or collecting personal data. Quantization may preserve
or alter the biases and unsafe behaviors of the underlying models; improved
quantization accuracy should not be interpreted as evidence of safety or
fairness. Downstream use should respect applicable model and dataset licenses
and include appropriate application-specific safety assessments.

\subsection*{Reproducibility statement}

\Cref{sec:method} describes the quantizer, alignment objective, structured
rotation, and deployment procedure.
\Cref{app:theoretical_foundations} provides the mathematical assumptions
and detailed proofs.
\Cref{sec:experiments} describes the evaluated models, calibration and
evaluation protocols, quantization settings, and metrics.
Additional metadata-budget and runtime analyses are provided in
\Cref{app:metadata_deployment}.
The implementation records experimental configurations and includes numerical
checks of rotation equivalence and structured-transform correctness.

\clearpage

\bibliography{iclr2027_conference}
\bibliographystyle{iclr2027_conference}

\clearpage

\appendix


\definecolor{PQTSpin}{HTML}{EEE7F4}
\definecolor{PQTPrefix}{HTML}{F8E9DD}

\section{Experimental Details and Additional Results}
\label{app:experiments}

\subsection{Experimental Protocol}
\label{app:setup}

\paragraph{Models.}
We evaluate Llama-3.2-3B, Llama-3.1-8B, and Llama-3.1-70B
(Section~\ref{subsec:dense}), the mixture-of-experts
Qwen3-30B-A3B-Base (Section~\ref{subsec:moe}), the Qwen3 Base family
at 0.6B, 1.7B, 4B, and 8B (Appendix~\ref{app:qwen3_dense}), and
Mistral-7B-v0.3 (Appendix~\ref{app:mistral}).
Every model runs in fp32 containers holding the released bf16
weights; the bf16 reference rows use the same containers, so
quantized and reference rows share one numerical path.

\vspace{-4pt}
\paragraph{Quantizers.}
Activations are dynamically quantized to asymmetric INT4 per-group
with $g=128$ at each linear input; each group stores an fp16 scale
and an fp16 zero-point for $4.25$ bits per value.
Weights are quantized with GPTQ to asymmetric INT4: per-group with
$g=128$ (fp16 scale and INT4 zero, $4.16$ bits) for Llama-3.2-3B,
the Qwen3 family, Qwen3-30B-A3B, and Mistral-7B, and per-channel
($4.00$ bits) for Llama-3.1-8B and Llama-3.1-70B.
On the 70B the Gram products of GPTQ are accumulated in fp64,
because the damped down-projection Hessian of its massive-activation
layer is indefinite in fp32 by an amount comparable to the damping.
The KV cache follows KIVI \citep{liu2024kivi}: keys are quantized
per channel in groups of 32 along the token axis and values per
token in groups of 128 along the head axis, both with fp16 metadata;
the most recent 32 tokens stay in full precision.
At context 2048 this gives $5.17$ bits per key and $4.43$ bits per
value.

\vspace{-4pt}
\paragraph{Rotations and calibration.}
Rotations act at the standard sites: $R_1$ on the residual stream,
folded into the embedding, the language-model head, and every linear
that reads from or writes to the residual after RMSNorm fusion;
$R_2$ on the value heads, folded into the value and output
projections; and $R_4$ online on the down-projection input.
The Hadamard baseline uses random-sign Hadamard matrices at all
sites; widths that are not powers of two (2560 and 9728) factor as
$128\times20$ and $128\times76$ with Paley blocks.
\pq{} replaces $R_1$ and $R_4$ by the rank-$k$ rotations of
Algorithm~\ref{alg:prismquant}, with $k=\max$ equal to the number of
group slots $n/128$ at each site, and uses a rank-one $R_2$.
Second moments are uncentered and estimated on 128 sequences of 2048
tokens from the WikiText-2 training split; the same sequences serve
as the GPTQ calibration set.
Before any quantizer is attached, every rotation reconstructs to a
relative round-trip error below $10^{-6}$ and the folded model
reproduces its bf16 reference to $|\Delta\mathrm{NLL}|\le10^{-5}$
per token.

\vspace{-4pt}
\paragraph{Evaluation.}
WikiText-2 perplexity is computed over 141 contiguous windows of
2048 test tokens with fp32 negative log-likelihoods, and C4
perplexity over 256 windows of 2048 tokens from the first validation
shard.
Zero-shot accuracy uses a pinned revision of lm-evaluation-harness
with \texttt{acc\_norm} on ARC-e, ARC-c, HellaSwag, OBQA, and PIQA
and \texttt{acc} on BoolQ, SIQA, and WinoGrande; the binomial
standard error of the eight-task mean is about $0.43$ points.
The Qwen3 family follows the Base-model evaluation of the Qwen3
technical report where the harness supports it: MMLU and MMLU-Redux
5-shot (the latter generative with an 8-token answer), GSM8K 4-shot
chain-of-thought with 512 new tokens and flexible answer extraction,
and ARC-e zero-shot.
End-to-end rows use three seeds; the activation-only ablations on
Llama average three paired rotation seeds, which vary the random
signs, the randomized eigensolver, and the GPTQ sample.

\vspace{-4pt}
\paragraph{Baselines.}
The main tables contain two kinds of rows.
Hadamard and \pq{} are run in our pipeline and share every setting
above except the rotation: the Hadamard row is the QuaRot
construction \citep{ashkboos2024quarot}, random-sign Hadamard
matrices at $R_1$, $R_2$, and $R_4$, placed under our group-128
asymmetric activation grid, and it can therefore exceed the published
QuaRot numbers, which quantize activations per token on a symmetric
grid.
The gain attributed to \pq{} is measured against the Hadamard row,
and the subspace-estimator comparison of Appendix~\ref{app:e27}
places the mechanism of the closest prior method inside the same
pipeline.

\vspace{-4pt}
\paragraph{Mixture of experts.}
Qwen3-30B-A3B-Base has 48 layers with 128 experts of intermediate
width 768, top-8 routing with renormalized weights, and no shared
expert.
$R_1$ is folded into the router's input columns, so routing logits
are unchanged in exact arithmetic; the router stays in bf16 and
reads the unquantized post-norm activation.
The expert input is quantized once per token in the $R_1$ basis and
shared by the eight routed experts.
Each expert receives its own $R_4$ at $k=6$, its full slot count,
and its own GPTQ Hessian; experts that see fewer than 512 routed
calibration tokens (1,346 of 6,144) fall back to the layer-pooled
Hessian rotated into the expert basis.
Cold-expert covariances use the shrinkage of
Eq.~\ref{eq:expert_covariance_shrinkage}.
With all rotations folded and no quantizer attached, perplexity moves
by less than $10^{-4}$ and top-8 routing decisions flip at most
$1.26\times$ as often as an identity probe run in fp32 ($1.78\times$
for Hadamard).
Effective widths are $4.16$ bits for weights, $4.25$ for activations,
and $5.17$/$4.43$ for keys/values; fused-kernel timing is not
reported for this model, since a routed-expert GEMM with per-expert
rotations requires its own kernel.

\subsection{Qwen3 Dense Family}
\label{app:qwen3_dense}

\begin{table}[!htbp]
\centering
\captionsetup{font=small,skip=4pt}

\caption{\textbf{W4A4KV4 quantization across the Qwen3 dense family.}
All models are Base checkpoints evaluated in our pipeline with three seeds.
WT2 and C4 report perplexity ($\downarrow$); task scores are
percentages ($\uparrow$).
MMLU and MMLU-R (Redux) are 5-shot; GSM8K is 4-shot CoT;
ARC-e is zero-shot\citep{raffel2020exploring,cobbe2021training,wang2024mmlu,gema2025we}.
Bold marks the best quantized value per model and column,
including rounded ties.}

\label{tab:qwen3_dense}

\begingroup
\fontsize{8pt}{9pt}\selectfont
\setlength{\tabcolsep}{2pt}
\renewcommand{\arraystretch}{1.06}
\setlength{\extrarowheight}{0.3pt}
\setlength{\aboverulesep}{1.8pt}
\setlength{\belowrulesep}{1.8pt}
\arrayrulecolor{black}

\begin{tabularx}{\linewidth}{
    l*{6}{>{\raggedleft\arraybackslash}X}
}
\toprule

& \multicolumn{2}{c}{\textbf{Perplexity} $\downarrow$}
& \multicolumn{4}{c}{\textbf{Task score} $\uparrow$} \\

\cmidrule(lr){2-3}
\cmidrule(lr){4-7}

\textbf{Method}
& WT2
& C4
& MMLU
& MMLU-R
& GSM8K
& ARC-e \\

\midrule


\rowcolor{PQTRef}
\textcolor{PQTInk}{\textbf{Qwen3-0.6B} (bf16)}
& 12.66 & 19.38 & 52.52 & 55.83 & 60.96 & 57.95 \\

Hadamard
& 16.50 & 25.36 & 43.46 & 43.49 & 28.96 & \pqtbest{56.69} \\

\pqtdashline

\rowcolor{PQTOurs}
\pqtours{} ($k=8$)
& \pqtbest{15.28} & \pqtbest{23.92} & 44.22 & \pqtbest{46.23}
& \pqtbest{31.54} & 53.62 \\

\rowcolor{PQTOursDeep}
\pqtours{} ($k=\max$)
& 15.38 & \pqtbest{23.92} & \pqtbest{44.69} & 45.27
& 28.66 & 56.36 \\

\midrule


\rowcolor{PQTRef}
\textcolor{PQTInk}{\textbf{Qwen3-1.7B} (bf16)}
& 9.40 & 15.02 & 62.68 & 66.98 & 72.78 & 68.48 \\

Hadamard
& 11.09 & 17.94 & 55.81 & 59.66 & 58.07 & 69.11 \\

\pqtdashline

\rowcolor{PQTOurs}
\pqtours{} ($k=8$)
& 10.33 & 16.53 & \pqtbest{58.20} & \pqtbest{62.05}
& \pqtbest{62.47} & 67.47 \\

\rowcolor{PQTOursDeep}
\pqtours{} ($k=\max$)
& \pqtbest{10.31} & \pqtbest{16.49} & 57.61 & 61.74
& 61.71 & \pqtbest{72.52} \\

\midrule


\rowcolor{PQTRef}
\textcolor{PQTInk}{\textbf{Qwen3-4B} (bf16)}
& 7.93 & 13.02 & 73.11 & 77.09 & 85.14 & 76.01 \\

Hadamard
& 8.76 & 14.40 & 69.23 & 73.55 & 74.83 & 74.75 \\

\pqtdashline

\rowcolor{PQTOurs}
\pqtours{} ($k=8$)
& \pqtbest{8.50} & 13.96 & \pqtbest{69.97} & 74.03
& \pqtbest{78.92} & \pqtbest{77.06} \\

\rowcolor{PQTOursDeep}
\pqtours{} ($k=\max$)
& 8.52 & \pqtbest{13.95} & 68.75 & \pqtbest{75.33}
& 78.70 & 74.75 \\

\midrule


\rowcolor{PQTRef}
\textcolor{PQTInk}{\textbf{Qwen3-8B} (bf16)}
& 8.80 & 11.66 & 76.83 & 81.35 & 87.11 & 80.01 \\

Hadamard
& 11.39 & 13.08 & 73.55 & 77.17 & 81.73 & 76.01 \\

\pqtdashline

\rowcolor{PQTOurs}
\pqtours{} ($k=8$)
& \pqtbest{9.73} & 12.31 & 74.39 & 78.52
& \pqtbest{84.31} & 80.68 \\

\rowcolor{PQTOursDeep}
\pqtours{} ($k=\max$)
& 9.85 & \pqtbest{12.28} & \pqtbest{74.58} & \pqtbest{79.51}
& 83.55 & \pqtbest{81.06} \\

\bottomrule
\end{tabularx}
\endgroup
\end{table}

Table~\ref{tab:qwen3_dense} runs the unchanged W4A4KV4 pipeline
across the Qwen3 Base family, so the \pq{}--Hadamard margin can be
read as a function of model size.
The smaller the model, the more W4A4KV4 costs and the more of that
cost \pq{} removes in absolute terms: Hadamard's WikiText-2
degradation falls from $30\%$ at 0.6B to $18\%$ at 1.7B and $10\%$
at 4B, the \pq{} margin from $1.22$ to $0.78$ and $0.26$ PPL, while
the fraction of the degradation removed stays between $31\%$ and
$46\%$.
Qwen3-8B breaks the trend: Hadamard degrades it by $29\%$, as much
as the 0.6B, and \pq{} removes $64\%$ of that ($11.39\to9.73$).
This is the model whose massive-activation layer places most of its
down-projection input on a group-constant direction, the direction
$R_4$ absorbs and a Hadamard spreads; the margin therefore tracks
how much of the Hadamard degradation is group-constant, not model
size per se.
Few-shot and generative metrics move with perplexity: at $k=8$,
\pq{} is above Hadamard on MMLU and GSM8K at every size, by
$0.7$--$2.4$ and $2.6$--$4.4$ points.
$k=8$ and $k=\max$ are within $0.12$ PPL of each other at every size
and trade places across benchmarks.

\subsection{Mistral-7B-v0.3}
\label{app:mistral}

\begin{table}[t]
\centering
\captionsetup{font=small,skip=4pt}

\caption{\textbf{W4A4KV4 quantization on Mistral-7B-v0.3.}
WT2/C4 perplexity ($\downarrow$) and zero-shot accuracy
(\%, $\uparrow$).
Avg.\,5 averages ARC-c, ARC-e, HellaS, PIQA, and WinoG;
Avg.\,8 covers all eight tasks.
The bf16 reference is ours; QuaRot and SmoothRot rows are their
reported RTN results, and cross-paper protocols differ.
A dash denotes an unreported result.
Bold marks the best displayed quantized result per column.}

\label{tab:mistral_sota}

\begingroup
\fontsize{8pt}{9pt}\selectfont
\setlength{\tabcolsep}{1.4pt}
\renewcommand{\arraystretch}{1.08}
\setlength{\extrarowheight}{0.3pt}

\setlength{\aboverulesep}{1.8pt}
\setlength{\belowrulesep}{1.8pt}
\arrayrulecolor{black}

\begin{tabularx}{\linewidth}{
    l*{12}{>{\raggedleft\arraybackslash}X}
}
\toprule

& \multicolumn{2}{c}{\textbf{PPL} $\downarrow$}
& \multicolumn{8}{c}{\textbf{Zero-shot accuracy} $\uparrow$}
& \multicolumn{2}{c}{\textbf{Mean} $\uparrow$} \\

\cmidrule(lr){2-3}
\cmidrule(lr){4-11}
\cmidrule(lr){12-13}

\textbf{Method}
& WT2
& C4
& ARC-c
& ARC-e
& BoolQ
& HellaS
& OBQA
& PIQA
& SIQA
& WinoG
& \mbox{Avg.\,5}
& \mbox{Avg.\,8} \\

\midrule

\rowcolor{PQTRef}
\textcolor{PQTInk}{\textbf{Mistral-7B-v0.3} (bf16)}
& 5.45 & 8.32
& 54.35 & 80.13 & 83.64 & 80.67
& 47.40 & 81.99 & 47.75 & 74.19
& 74.27 & 68.77 \\

QuaRot\pqtvenue{NeurIPS}{24}
& 6.41 & 10.01
& 45.22 & 72.26 & --- & 75.14
& --- & 79.22 & --- & 68.82
& 68.13 & --- \\

SmoothRot\pqtvenue{SMC}{25}
& 6.19 & 9.78
& 46.59 & 73.19 & --- & 75.97
& --- & 79.11 & --- & 68.11
& 68.59 & --- \\

\rowcolor{PQTSpin}
SpinQuant\pqtvenue{ICLR}{25}
& 5.80
& ---
& 52.50
& 77.30
& 80.20
& 79.20
& \pqtbest{58.40}
& 80.30
& \pqtbest{48.90}
& 72.30
& 72.32
& \pqtbest{68.60} \\

\rowcolor{PQTPrefix}
PrefixQuant\pqtvenue{TPAMI}{26}
& 5.76 & ---
& 50.00 & 77.74 & --- & 78.39
& --- & 80.41 & --- & 70.88
& 71.48 & --- \\

\pqtdashline

Hadamard
& 5.64
& 8.65
& \pqtbest{52.82}
& \pqtbest{79.00}
& 81.96
& \pqtbest{79.84}
& 45.00
& 80.96
& 46.32
& 72.45
& 73.01
& 67.29 \\

\rowcolor{PQTOurs}
\pqtours{} ($k=8$)
& \pqtbest{5.62}
& 8.61
& 52.73
& 78.83
& \pqtbest{82.32}
& 79.39
& 46.40
& \pqtbest{81.94}
& 47.24
& \pqtbest{72.53}
& \pqtbest{73.08}
& 67.67 \\

\rowcolor{PQTOursDeep}
\pqtours{} ($k=\max$)
& 5.63
& \pqtbest{8.59}
& 52.47
& 78.96
& 81.87
& 79.27
& 45.20
& 81.01
& 46.57
& 72.06
& 72.75
& 67.18 \\

\bottomrule
\end{tabularx}
\endgroup
\end{table}

Mistral-7B-v0.3 is a model that W4A4KV4 damages little: in
Table~\ref{tab:mistral_sota}, Hadamard sits $0.19$ PPL above bf16
and loses $1.48$ eight-task points.
\pq{} at $k=8$ lowers WikiText-2 perplexity to $5.62$ and raises the
eight-task mean by $0.38$ points, recovering $26\%$ of that loss;
$k=\max$ gives the lowest C4 perplexity ($8.59$).
The accuracy margin is inside one standard error of the eight-task
mean and is reported as such.
Against published rows, $5.62$ is below every displayed entry
(PrefixQuant $5.76$, SpinQuant $5.80$, SmoothRot $6.19$, QuaRot
$6.41$) and also below the GPTQ variants that QuaRot and SmoothRot
report for perplexity only ($5.79$ and $5.81$)
\citep{ashkboos2024quarot,czako2025smoothrot,liu2025spinquant,chen2026prefixquant},
while SpinQuant retains the higher reported eight-task mean
($68.60\%$).

\subsection{Additional Experiments}
\label{app:additional_experiments}
\label{app:e27} 

This section reports the three ablations summarized in
Section~\ref{subsec:ablations}: the allocation of activation metadata
and the calibration of sparsely routed experts
(Table~\ref{tab:app_ablations}), and the token population that
defines the subspace (Table~\ref{tab:e27_tokens}).
Panel~(a) of Table~\ref{tab:app_ablations} is an activation-only
diagnostic with bf16 weights and KV states; panel~(b) is the full
W4A4KV4 MoE setting, so absolute perplexities are not comparable
across panels.

\begin{table}[t]
\centering
\captionsetup{font=small,skip=4pt}

\caption{\textbf{Ablations of metadata allocation and expert calibration.}
\textbf{(a)}~Activation-only WikiText-2 PPL ($\downarrow$) under
matched representation budgets; H/PQ denote Hadamard/\pq{} at
$k=\max$, and \textcolor{PQTGreen}{green parentheses} give
$\mathrm{PQ}-\mathrm{H}$.
\textbf{(b)}~W4A4KV4 WikiText-2 PPL ($\downarrow$) for the MoE
calibration variants at $k=\max$; $\Delta$PPL is relative to
shrinkage, from unrounded values.
Bold denotes the lower PPL within each H/PQ pair in (a) and the
lowest PPL in (b); \colorbox{PQTMint}{mint rows} mark the
configuration adopted in the main tables.}

\label{tab:app_ablations}

\begingroup
\fontsize{8pt}{9.5pt}\selectfont
\setlength{\tabcolsep}{3pt}
\renewcommand{\arraystretch}{1.08}
\setlength{\extrarowheight}{0.3pt}
\setlength{\aboverulesep}{2pt}
\setlength{\belowrulesep}{2pt}
\arrayrulecolor{black}

\begin{tabularx}{\linewidth}{
    >{\raggedright\arraybackslash}p{0.36\linewidth}
    *{4}{>{\raggedleft\arraybackslash}X}
}
\toprule


\rowcolor{PQTRef}
\multicolumn{5}{l}{
    \textcolor{PQTInk}{
        \textbf{(a) Activation metadata allocation: WikiText-2 PPL}
    }
} \\

\rowcolor{PQTBand}
\textbf{Configuration}
& \multicolumn{2}{c}{\textbf{Llama-3.2-3B}}
& \multicolumn{2}{c}{\textbf{Llama-3.1-8B}} \\

\cmidrule(lr){2-3}
\cmidrule(lr){4-5}

\rowcolor{PQTBand}
$(g,m)$; bits/value
& H
& \textcolor{PQTInk}{\textbf{PQ}}
& H
& \textcolor{PQTInk}{\textbf{PQ}} \\

\midrule

$(256,1)$; 4.13
& 7.79
& \pqtbest{7.74}\,\pqtdelta{-0.05}
& 6.37
& \pqtbest{6.31}\,\pqtdelta{-0.06} \\

$(256,2)$; 4.19
& 7.79
& \pqtbest{7.73}\,\pqtdelta{-0.06}
& 6.38
& \pqtbest{6.30}\,\pqtdelta{-0.08} \\

$(256,3)$; 4.25
& 7.79
& \pqtbest{7.73}\,\pqtdelta{-0.06}
& 6.37
& \pqtbest{6.30}\,\pqtdelta{-0.07} \\

\rowcolor{PQTMint}
$(128,1)$; 4.25 \pqtdefault{}
& 7.77
& \pqtbest{7.71}\,\pqtdelta{-0.06}
& 6.35
& \pqtbest{6.29}\,\pqtdelta{-0.06} \\

$(128,2)$; 4.38
& 7.77
& \pqtbest{7.70}\,\pqtdelta{-0.07}
& 6.35
& \pqtbest{6.29}\,\pqtdelta{-0.06} \\

$(64,1)$; 4.50
& 7.74
& \pqtbest{7.68}\,\pqtdelta{-0.06}
& 6.34
& \pqtbest{6.26}\,\pqtdelta{-0.08} \\

\midrule


\rowcolor{PQTRef}
\multicolumn{5}{l}{
    \textcolor{PQTInk}{
        \textbf{(b) MoE calibration: Qwen3-30B-A3B, W4A4KV4}
    }
} \\

\rowcolor{PQTBand}
\multicolumn{3}{l}{\textbf{Calibration scheme}}
& \textbf{PPL} $\downarrow$
& \mbox{$\Delta$\textbf{PPL}} \\

\midrule

\multicolumn{3}{l}{
    Per-expert covariance; cold experts use Hadamard
}
& 6.43
& \pqtworse{+0.01} \\

\multicolumn{3}{l}{
    Pooled covariance for all experts
}
& 6.45
& \pqtworse{+0.02} \\

\rowcolor{PQTMint}
\multicolumn{3}{l}{
    \pqtours{} with covariance shrinkage \pqtdefault{}
}
& \pqtbest{6.42}
& \pqtzero{0.00} \\

\bottomrule
\end{tabularx}
\endgroup
\end{table}

\paragraph{Metadata allocation.}
Alignment is worth more than the metadata separating our coarsest and
finest quantizers, and scale resolution and represented directions
are distinct levers.
Table~\ref{tab:app_ablations}a varies the group size $g$ and the
number $m$ of represented within-group directions, including the
constant one, at a budget of $4+16(m+1)/g$ bits per value; every
\pq{} row uses $k=\max$ and three paired seeds.
\pq{} is lower at every matched configuration; at $(256,1)$,
$4.13$ bits, it already matches or beats Hadamard at $(64,1)$,
$4.50$ bits ($7.74$ vs.\ $7.74$; $6.31$ vs.\ $6.34$).
Extra directions help only when energy is aligned into them:
Hadamard is flat across $m$ at both group sizes, whereas \pq{} gains
up to $0.01$ PPL.
At the shared $4.25$-bit budget, the default $(128,1)$ beats
$(256,3)$ on both models, so halving the group is worth more than
two additional directions.

\paragraph{Calibrating sparsely routed experts.}
The gain does not hinge on how cold experts are calibrated.
Over the calibration set, $2{,}002$ of the $6{,}144$ experts receive
fewer than $2{,}048$ routed tokens, so we shrink each expert's
second moment toward its layer's pooled estimate,
\begin{equation}
    \widetilde{\Sigma}_e
    =
    \frac{n_e\Sigma_e+n_0\Sigma_{\mathrm{pool}}}{n_e+n_0},
    \qquad n_0=2048,
    \label{eq:expert_covariance_shrinkage}
\end{equation}
where $n_e$ is the expert's routed-token count; the pooled term
dominates exactly for the cold experts.
Table~\ref{tab:app_ablations}b compares this with a plain Hadamard
$R_4$ for the cold experts and with one pooled covariance for all
experts.
Shrinkage is best, but the three schemes differ by at most $0.023$
PPL, an order of magnitude less than the $0.26$ margin over
Hadamard: alignment on the well-populated experts carries the gain.

\paragraph{Which tokens define the subspace.}
\pq{} estimates its subspace from the uncentered second moment of all
calibration tokens, on the premise that the quantizer's free directions
are fixed and the only question is where the activation energy lies.
A natural alternative is to let extreme tokens define the subspace, as
outlier-driven methods do; OffQ \citep{wang2026offq}, for instance, keeps
only the maximum-$L_\infty$ token of each calibration sequence.
This pre-registered ablation tests that choice inside the \pq{} pipeline
with everything else fixed, so the token population is the only variable.
Variant~A is our estimator; B keeps the top-1 token per sequence
(OffQ-style); C keeps the global top $1\%$ of tokens per layer and site;
both B and C are also run with BOS (position~0) removed from the candidate
set.
All variants share the Householder construction, the slot
permutations of~A, the block Hadamard, $k=\max$, the three paired
rotation seeds, the 64 evaluation chunks, and the activation-only
protocol of Table~\ref{tab:rank_ablation}.
B uses a thin SVD of the selected tokens; where fewer than eight
directions are identified (7 layer/site estimates on 3B, 5 on 8B)
the remainder is completed by a seeded random orthogonal complement.
Table~\ref{tab:e27_tokens} reports the results.

\begin{table}[t]
\centering
\captionsetup{font=small,skip=4pt}

\caption{\textbf{Which tokens define the subspace.}
Activation-only WikiText-2 PPL with bf16 weights and KV states,
three paired seeds, $k=\max$.
\textbf{(a)}~Both sites quantized; $\Delta$ is the paired
difference to the all-token estimator with its 90\% Student-$t$
interval.
\textbf{(b)}~$\Delta$ when only one site is quantized (BOS
included); bold marks intervals that exclude zero.
\textbf{(c)}~Diagnostics at the down-projection input: captured
group-constant energy $f$ and per-group range relative to Hadamard.
Mint rows mark our estimator.}

\label{tab:e27_tokens}

\begingroup
\fontsize{8pt}{9.5pt}\selectfont
\setlength{\tabcolsep}{3pt}
\renewcommand{\arraystretch}{1.08}
\setlength{\extrarowheight}{0.3pt}
\setlength{\aboverulesep}{2pt}
\setlength{\belowrulesep}{2pt}
\arrayrulecolor{black}

\begin{tabularx}{\linewidth}{
    >{\raggedright\arraybackslash}p{0.27\linewidth}
    >{\raggedleft\arraybackslash}p{0.07\linewidth}
    >{\raggedleft\arraybackslash}X
    >{\raggedleft\arraybackslash}p{0.07\linewidth}
    >{\raggedleft\arraybackslash}X
}
\toprule

\rowcolor{PQTRef}
\multicolumn{5}{l}{
    \textcolor{PQTInk}{\textbf{(a) Both sites quantized: WikiText-2 PPL}}
} \\

\rowcolor{PQTBand}
\textbf{Subspace estimator}
& \multicolumn{2}{c}{\textbf{Llama-3.2-3B}}
& \multicolumn{2}{c}{\textbf{Llama-3.1-8B}} \\

\cmidrule(lr){2-3}
\cmidrule(lr){4-5}

\rowcolor{PQTBand}
& PPL & $\Delta$ [90\% CI]
& PPL & $\Delta$ [90\% CI] \\

\midrule

\rowcolor{PQTMint}
All tokens (\pq{})
& \pqtbest{7.705} & \pqtzero{0}
& \pqtbest{6.285} & \pqtzero{0} \\

Top-1 token / seq.\ (OffQ-style)
& 7.723 & \pqtworse{+0.017}\,\mbox{\scriptsize[$-$0.002, $+$0.036]}
& 6.301 & \pqtworse{+0.016}\,\mbox{\scriptsize[$+$0.012, $+$0.020]} \\

\quad BOS excluded
& 7.800 & \pqtworse{+0.095}\,\mbox{\scriptsize[$+$0.092, $+$0.098]}
& 6.342 & \pqtworse{+0.057}\,\mbox{\scriptsize[$+$0.049, $+$0.065]} \\

Global top 1\% of tokens
& 7.711 & \pqtworse{+0.006}\,\mbox{\scriptsize[$-$0.012, $+$0.024]}
& 6.293 & \pqtworse{+0.008}\,\mbox{\scriptsize[$+$0.002, $+$0.014]} \\

\quad BOS excluded
& 7.806 & \pqtworse{+0.101}\,\mbox{\scriptsize[$+$0.072, $+$0.129]}
& 6.329 & \pqtworse{+0.044}\,\mbox{\scriptsize[$+$0.037, $+$0.052]} \\

\pqtdashline

Hadamard (no alignment)
& 7.771 & \pqtworse{+0.066}\,\mbox{\scriptsize[$+$0.039, $+$0.093]}
& 6.346 & \pqtworse{+0.061}\,\mbox{\scriptsize[$+$0.056, $+$0.066]} \\

\midrule

\rowcolor{PQTRef}
\multicolumn{5}{l}{
    \textcolor{PQTInk}{\textbf{(b) One site quantized: $\Delta$ PPL vs.\ all tokens}}
} \\

\rowcolor{PQTBand}
\textbf{Subspace estimator}
& \multicolumn{2}{c}{\textbf{Llama-3.2-3B}}
& \multicolumn{2}{c}{\textbf{Llama-3.1-8B}} \\

\cmidrule(lr){2-3}
\cmidrule(lr){4-5}

\rowcolor{PQTBand}
& q/k/v & down
& q/k/v & down \\

\midrule

Top-1 token / seq.\ (OffQ-style)
& +0.002 & \pqtbest{+0.012}
& \pqtbest{+0.003} & \pqtbest{+0.017} \\

Global top 1\% of tokens
& +0.002 & \pqtbest{+0.013}
& \pqtbest{+0.003} & \pqtbest{+0.009} \\

\pqtdashline

Hadamard (no alignment)
& \pqtbest{+0.017} & \pqtbest{+0.037}
& \pqtbest{+0.018} & \pqtbest{+0.053} \\

\midrule

\rowcolor{PQTRef}
\multicolumn{5}{l}{
    \textcolor{PQTInk}{\textbf{(c) Down-projection input: captured energy and range}}
} \\

\rowcolor{PQTBand}
\textbf{Subspace estimator}
& \multicolumn{2}{c}{\textbf{Llama-3.2-3B}}
& \multicolumn{2}{c}{\textbf{Llama-3.1-8B}} \\

\cmidrule(lr){2-3}
\cmidrule(lr){4-5}

\rowcolor{PQTBand}
& $f$ $\uparrow$ & range/H $\downarrow$
& $f$ $\uparrow$ & range/H $\downarrow$ \\

\midrule

\rowcolor{PQTMint}
All tokens (\pq{})
& \pqtbest{0.333} & \pqtbest{0.747}
& \pqtbest{0.324} & \pqtbest{0.746} \\

Top-1 token / seq.\ (OffQ-style)
& 0.154 & 0.845
& 0.152 & 0.839 \\

Global top 1\% of tokens
& 0.226 & 0.812
& 0.215 & 0.811 \\

\pqtdashline

Hadamard (no alignment)
& 0.007 & 1.000
& 0.007 & 1.000 \\

\bottomrule
\end{tabularx}
\endgroup
\end{table}

Three findings follow.
First, the all-token estimator has the lowest perplexity on both
models at every site.
With both sites quantized, top-1 selection costs $+0.016$ PPL on
Llama-3.1-8B with an interval that excludes zero and $+0.017$ on
Llama-3.2-3B, where the interval overlaps zero at both sites but
excludes it at the down-projection input alone ($+0.012$
$[+0.003,+0.022]$); the global top-$1\%$ variant sits between the
two.
Panel~(b) localizes the effect: at the q/k/v input the estimators are
within $0.003$ PPL of each other, at the down-projection input they
separate.
Second, the outlier-token estimator relies on the BOS token.
Removing BOS from the candidate set costs a further $0.04$--$0.08$
PPL and lands at or above Hadamard's perplexity on both models, and
with BOS included the selected token set is rank deficient at
several layers.
Third, the diagnostics in panel~(c) explain why the perplexity gap
is nevertheless modest.
The top-1 subspace captures less than half the group-constant energy
of ours at the down-projection input ($f=0.15$ vs.\ $0.33$), reduces
the per-group range less ($0.84$ vs.\ $0.75$ of Hadamard's), and its
leading eight directions lie $77^\circ$ from ours in principal
angle; yet it still recovers about three quarters of the all-token
gain over Hadamard on both models.
A few dominant directions carry most of the benefit of alignment,
and pooling all tokens supplies the remainder.

\paragraph{Calibration and estimation robustness.}
\label{app:robustness}
The gain depends on neither the calibration set nor the accuracy of the
eigensolver, only on the captured energy $f$.
Table~\ref{tab:app_robustness}a varies the calibration set for
Qwen3-4B-Base at $k=\max$: shrinking or doubling the 128 WikiText-2
sequences moves the WikiText-2 gain by less than $7\%$, and calibrating
on C4 instead retains $95\%$ of it while matching the WikiText-2
calibration on C4 evaluation.
The corpus change reassigns $99\%$ of the residual coordinates and
lowers the down-projection $f$ from $0.315$ to $0.273$, yet perplexity
barely moves, so the greedy placement is not load-bearing and modest
losses of captured energy are tolerated.
Panel~(b) varies the randomized eigensolver: a single pass captures
almost nothing ($f=0.009$) and costs $0.059$ PPL, two passes recover
$f=0.290$, and the default three passes are within seed noise of an
exact fp64 eigendecomposition on both models.
Principal angles between the randomized and exact subspaces stay near
$90^\circ$ throughout, because the trailing directions of a
$76$-dimensional leading eigenspace are nearly degenerate; perplexity
nevertheless tracks $f$ monotonically, which is the quantity the
objective controls.

\providecommand{\pqtgain}[1]{\textcolor{PQTGreen}{$#1$}}
\providecommand{\pqtshare}[1]{\textcolor{PQTInk}{$#1$}}

\begin{table}[t]
\centering
\captionsetup{font=small,skip=4pt}
\caption{\textbf{Calibration and estimation robustness on Qwen3-4B-Base.}
Activation-only WikiText-2 (WT2) and C4 PPL, three paired seeds, $k=\max$;
$\Delta$ is the paired difference to Hadamard with its 90\% interval.
\textbf{(a)}~Calibration set size and corpus.
\textbf{(b)}~Randomized eigensolver passes versus exact fp64
eigendecomposition; $f$ is the captured energy at the down-projection
input, $\Delta$ is relative to the exact solver.
Mint rows mark the default.}
\label{tab:app_robustness}
\begingroup
\fontsize{7pt}{8pt}\selectfont
\setlength{\tabcolsep}{1.8pt}
\renewcommand{\arraystretch}{1.00}
\setlength{\aboverulesep}{1.8pt}
\setlength{\belowrulesep}{1.8pt}
\arrayrulecolor{black}
\begin{tabularx}{\linewidth}{
    >{\raggedright\arraybackslash}p{0.26\linewidth}
    >{\raggedleft\arraybackslash}p{0.08\linewidth}
    >{\raggedleft\arraybackslash}X
    >{\raggedleft\arraybackslash}p{0.08\linewidth}
    >{\raggedleft\arraybackslash}X
}
\toprule
\rowcolor{PQTRef}
\multicolumn{5}{l}{\textcolor{PQTInk}{\textbf{(a) Calibration set: size and corpus}}} \\
\rowcolor{PQTBand}
\textbf{Calibration} & WT2 PPL & $\Delta$ [90\% CI] & C4 PPL & $\Delta$ [90\% CI] \\
\midrule
Hadamard (no calibration)
& 7.877 & \pqtzero{0}
& 13.160 & \pqtzero{0} \\
\pqtdashline
WT2, 64 seq.
& 7.690 & \pqtgain{-0.187}\,\mbox{\scriptsize[$-$0.199, $-$0.174]}
& 12.851 & \pqtgain{-0.309}\,\mbox{\scriptsize[$-$0.330, $-$0.289]} \\
\rowcolor{PQTMint}
WT2, 128 seq.\ \pqtdefault{}
& 7.694 & \pqtgain{-0.183}\,\mbox{\scriptsize[$-$0.194, $-$0.171]}
& 12.844 & \pqtgain{-0.316}\,\mbox{\scriptsize[$-$0.337, $-$0.295]} \\
WT2, 256 seq.
& 7.703 & \pqtgain{-0.174}\,\mbox{\scriptsize[$-$0.185, $-$0.163]}
& 12.852 & \pqtgain{-0.309}\,\mbox{\scriptsize[$-$0.330, $-$0.287]} \\
C4, 128 seq.
& 7.703 & \pqtgain{-0.174}\,\mbox{\scriptsize[$-$0.185, $-$0.162]}
& 12.852 & \pqtgain{-0.309}\,\mbox{\scriptsize[$-$0.330, $-$0.287]} \\
\midrule
\rowcolor{PQTRef}
\multicolumn{5}{l}{\textcolor{PQTInk}{\textbf{(b) Eigensolver accuracy}}} \\
\rowcolor{PQTBand}
\textbf{Solver} & Ritz res. & $f$ (down) & WT2 PPL & $\Delta$ vs.\ exact [90\% CI] \\
\midrule
Randomized, 1 pass
& 2.8 & 0.009 & 7.752
& \pqtworse{+0.059}\,\mbox{\scriptsize[$+$0.051, $+$0.067]} \\
Randomized, 2 passes
& 0.43 & 0.290 & 7.700
& \pqtworse{+0.007}\,\mbox{\scriptsize[$+$0.000, $+$0.013]} \\
\rowcolor{PQTMint}
Randomized, 3 passes \pqtdefault{}
& 0.12 & 0.325 & 7.697
& \pqtworse{+0.004}\,\mbox{\scriptsize[$-$0.002, $+$0.011]} \\
Exact (fp64)
& $<\!10^{-14}$ & 0.334 & \pqtbest{7.693} & \pqtzero{0} \\
\pqtdashline
\multicolumn{5}{l}{\textcolor{PQTMuted}{Llama-3.2-3B: 3 passes 7.705 ($f=0.314$) vs.\ exact 7.700 ($f=0.322$), $\Delta=+0.005$ [$-$0.000, $+$0.010]}} \\
\bottomrule
\end{tabularx}
\endgroup
\end{table}

\paragraph{Does the gain need the offset?}
\label{app:offset}
At $g=128$ the gain does not require the affine offset; the offset
is what makes the aligned level free, but flatness and localization
carry most of the perplexity.
Table~\ref{tab:app_offset}a swaps the quantizer while keeping the
rotation bit-identical: under symmetric per-group INT4, where no
offset exists, \pq{}'s margin over Hadamard is unchanged on both
models ($-0.194$ vs.\ $-0.179$ on Qwen3-4B-Base, $-0.060$ vs.\
$-0.061$ on Llama-3.2-3B).
The reason is what the rotation does to the leading directions
themselves: a dense Hadamard turns a distributed eigenvector into a
Gaussian-like vector whose largest coordinate is several times its
r.m.s., whereas \pq{} makes it exactly uniform within one group and
zero elsewhere, so any quantizer that scales by a group extremum sees a
smaller range whether or not it can absorb the level.
Panel~(b) isolates the offset inside the asymmetric format by moving
the anchors to a nonconstant Walsh column of the same group, which
keeps the aligned level flat and confined to one group but exposes it
to the range.
The cost is $16$--$25\%$ of the total gain, and the same move is
invisible under symmetric quantization, where the level was never
absorbed; the anchor groups' measured range grows by only $11$--$19\%$,
because the level and the residual extremes do not add linearly.
The offset dominates when a group is the whole token: with one offset
per token ($g=d$, $k=1$) it accounts for $37\%$ (Qwen) and $69\%$
(Llama) of the gain.
Finally, quantizing only the $0.15\%$ of positions flagged as BOS or
massive (top $0.1\%$ residual-stream $L_\infty$ at the layer with the
largest group-constant share) yields $20\%$ and $26\%$ of the full gain
on the two models, and quantizing the remaining $99.85\%$ yields the
rest, with the two parts adding to the all-token result within
$0.012$ PPL; the tokens with the largest crest factor benefit most, as
flatness predicts, while the offset-specific part of the gain is not
concentrated on them.
The split also shows how \pq{} interacts with prefix-based outlier
handling \citep{chen2026prefixquant}: with BOS and massive positions
kept in full precision, an idealized form of that approach, \pq{}
retains $72$--$74\%$ of its margin over Hadamard, so the two are
largely complementary.
Together with the subspace-estimator ablation above, this fixes the
reading of the objective: the group-constant directions are the ones a
per-group quantizer tolerates best, and the offset is the third of three
reasons.

\begin{table}[t]
\centering
\captionsetup{font=small,skip=4pt}
\caption{\textbf{Quantizer format and target column.}
Activation-only WikiText-2 PPL, three paired seeds, both sites quantized.
\textbf{(a)}~The same \pq{} rotation ($k=\max$; $k=1$ for $g=d$) under
four quantizers; $\Delta$ is PQ minus Hadamard, and every 90\% interval
excludes zero.
\textbf{(b)}~Anchors moved from the constant Walsh column to a
nonconstant one at $g=128$, $k=\max$; $\Delta$ is relative to the
constant column under the asymmetric and symmetric formats, and $s$ is
the offset-specific share
$\Delta_{\rm asym}/(\mathrm{PPL}_{\rm Had}-\mathrm{PPL}_{\rm PQ})$.
Mint rows mark the configuration of the main tables.}
\label{tab:app_offset}
\begingroup
\fontsize{7pt}{8pt}\selectfont
\setlength{\tabcolsep}{1.8pt}
\renewcommand{\arraystretch}{1.00}
\setlength{\aboverulesep}{1.8pt}
\setlength{\belowrulesep}{1.8pt}
\arrayrulecolor{black}
\begin{tabularx}{\linewidth}{
    >{\raggedright\arraybackslash}p{0.30\linewidth}
    *{6}{>{\raggedleft\arraybackslash}X}
}
\toprule
\rowcolor{PQTRef}
\multicolumn{7}{l}{\textcolor{PQTInk}{\textbf{(a) Quantizer format}}} \\
\rowcolor{PQTBand}
\textbf{Quantizer}
& \multicolumn{3}{c}{\textbf{Qwen3-4B-Base}}
& \multicolumn{3}{c}{\textbf{Llama-3.2-3B}} \\
\cmidrule(lr){2-4}\cmidrule(lr){5-7}
\rowcolor{PQTBand}
& Had & PQ & $\Delta$ & Had & PQ & $\Delta$ \\
\midrule
\rowcolor{PQTMint}
Asymmetric, $g=128$ \pqtdefault{}
& 7.877 & \pqtbest{7.697} & \pqtgain{-0.179}
& 7.767 & \pqtbest{7.705} & \pqtgain{-0.061} \\
Symmetric, $g=128$
& 7.958 & \pqtbest{7.764} & \pqtgain{-0.194}
& 7.819 & \pqtbest{7.759} & \pqtgain{-0.060} \\
Asymmetric, $g=d$ ($k=1$)
& 8.075 & \pqtbest{7.857} & \pqtgain{-0.218}
& 7.926 & \pqtbest{7.886} & \pqtgain{-0.041} \\
Symmetric, $g=d$ ($k=1$)
& 8.178 & \pqtbest{8.041} & \pqtgain{-0.137}
& 8.008 & \pqtbest{7.995} & \pqtgain{-0.013} \\
\midrule
\rowcolor{PQTRef}
\multicolumn{7}{l}{\textcolor{PQTInk}{\textbf{(b) Target column at $g=128$, $k=\max$}}} \\
\rowcolor{PQTBand}
\textbf{Anchor target}
& \multicolumn{3}{c}{\textbf{Qwen3-4B-Base}}
& \multicolumn{3}{c}{\textbf{Llama-3.2-3B}} \\
\cmidrule(lr){2-4}\cmidrule(lr){5-7}
\rowcolor{PQTBand}
& $\Delta_{\rm asym}$ & $\Delta_{\rm sym}$ & $s$
& $\Delta_{\rm asym}$ & $\Delta_{\rm sym}$ & $s$ \\
\midrule
\rowcolor{PQTMint}
Constant column \pqtdefault{}
& \pqtzero{0} & \pqtzero{0} & ---
& \pqtzero{0} & \pqtzero{0} & --- \\
Sequency-1 column ($\pm$ halves)
& \pqtworse{+0.029} & +0.009 & \pqtshare{0.16}
& \pqtworse{+0.015} & +0.005 & \pqtshare{0.25} \\
Alternating column
& \pqtworse{+0.023} & +0.001 & \pqtshare{0.13}
& \pqtworse{+0.013} & +0.004 & \pqtshare{0.22} \\
\bottomrule
\end{tabularx}
\endgroup
\end{table}

\paragraph{Construction ablations.}
\label{app:construction}
The signed permutation is not load-bearing for perplexity.
Table~\ref{tab:app_construction} varies it on Qwen3-4B-Base with the
Householder factor $G$ fixed.
Energy balancing does what it is designed to do, equalizing residual
energy across groups (max/median spread $1.0$ versus $3.4$ for the
identity permutation at the down-projection input), yet identity and
random permutations reach the same perplexity within seed noise at both
ranks: per-group scales already absorb the imbalance that balancing
removes.
At $k=8$, filling the unused constant slots with the highest-energy
coordinates instead of the lowest-energy ones improves perplexity by
$0.009$, because those slots are as tolerant as the anchored ones
(Appendix~\ref{app:offset}); we keep the low-energy fillers in the main
tables so that the rank sweep of Table~\ref{tab:rank_ablation} measures
eigenspace alignment alone, and note the improvement as available at
reduced rank.
Random signs matter only at reduced rank: fixing $D=+1$ costs $0.017$ at
$k=8$ and nothing at $k=\max$, where every constant slot is eigen-aligned
and the sign pattern no longer correlates the fillers across groups.

\begin{table}[t]
\centering
\captionsetup{font=small,skip=4pt}
\caption{\textbf{Anchor placement, fillers, and signs on Qwen3-4B-Base.}
Activation-only WikiText-2 PPL, three paired seeds, $G$ identical across
rows; $\Delta$ is relative to the default at the same rank with its 90\%
interval. Hadamard: $7.877$.}
\label{tab:app_construction}
\begingroup
\fontsize{7pt}{8pt}\selectfont
\setlength{\tabcolsep}{1.8pt}
\renewcommand{\arraystretch}{1.00}
\setlength{\aboverulesep}{1.8pt}
\setlength{\belowrulesep}{1.8pt}
\arrayrulecolor{black}
\begin{tabularx}{\linewidth}{
    >{\raggedright\arraybackslash}p{0.34\linewidth}
    >{\raggedleft\arraybackslash}p{0.08\linewidth}
    >{\raggedleft\arraybackslash}X
    >{\raggedleft\arraybackslash}p{0.08\linewidth}
    >{\raggedleft\arraybackslash}X
}
\toprule
\rowcolor{PQTBand}
\textbf{Variant}
& \multicolumn{2}{c}{$k=8$} & \multicolumn{2}{c}{$k=\max$} \\
\cmidrule(lr){2-3}\cmidrule(lr){4-5}
\rowcolor{PQTBand}
& PPL & $\Delta$ [90\% CI] & PPL & $\Delta$ [90\% CI] \\
\midrule
\rowcolor{PQTMint}
Balanced $\Pi$, low-energy fillers \pqtdefault{}
& 7.707 & \pqtzero{0}
& 7.697 & \pqtzero{0} \\
Identity $\Pi$
& 7.710 & +0.002\,\mbox{\scriptsize[$-$0.004, $+$0.009]}
& 7.697 & $-$0.001\,\mbox{\scriptsize[$-$0.007, $+$0.006]} \\
Random $\Pi$
& 7.708 & +0.001\,\mbox{\scriptsize[$-$0.005, $+$0.007]}
& 7.694 & $-$0.003\,\mbox{\scriptsize[$-$0.011, $+$0.004]} \\
High-energy fillers
& \pqtbest{7.698} & \pqtgain{-0.009}\,\mbox{\scriptsize[$-$0.015, $-$0.003]}
& 7.697 & \pqtzero{0}\,\mbox{\scriptsize(identical)} \\
Fixed signs $D=+1$
& 7.724 & \pqtworse{+0.017}\,\mbox{\scriptsize[$+$0.011, $+$0.023]}
& 7.697 & $-$0.001\,\mbox{\scriptsize[$-$0.007, $+$0.005]} \\
\bottomrule
\end{tabularx}
\endgroup
\end{table}

\subsection{Algorithm}
\label{app:algorithm}

Algorithm~\ref{alg:prismquant} lists the complete procedure.
Stage~I computes, for every site $s$, the null-space-aligned rotation
from uncentered second moments: the top-$k_s$ eigendirections, with
$k_s$ the site's rank capped by its number of group slots, are mapped
onto the first coordinate of $k_s$ distinct groups by a sequence of
Householder reflections held in compact-WY form; a signed permutation
then places the remaining coordinates, and the block Hadamard turns
each anchored coordinate into that group's constant direction.
Stage~II folds $R_1$ and $R_2$ into the surrounding weights and
registers the online $R_4$ as a rank-$k_s$ correction inside the
down-projection input; the post-RoPE query/key rotation of QuaRot,
$R_3$, is the identity here because keys are quantized per channel
without rotation (\Cref{sec:deployment}), and it is listed only so that
the four-site convention is complete.
Stage~III quantizes the rotated weights with GPTQ and attaches the
activation and KV quantizers; Stage~IV is the grouped asymmetric INT4
quantizer, whose fp16 offset represents the group-constant level.

\begin{algorithm}[!htbp]
\caption{%
  \textcolor{PQTInk}{%
    \textbf{PrismQuant: structured alignment and grouped INT4}%
  }%
}
\label{alg:prismquant}

\fontsize{9pt}{10.0pt}\selectfont

\begingroup
\algrenewcommand\algorithmicrequire{%
  \textcolor{PQTInk}{\textbf{Input:}}%
}
\algrenewcommand\algorithmicensure{%
  \textcolor{PQTInk}{\textbf{Output:}}%
}
\algrenewcommand\alglinenumber[1]{%
  \scriptsize\textcolor{PQTMuted}{#1}%
}

\begin{algorithmic}[1]

\Require Model $\mathcal M$ with $L$ layers; calibration data
         $\mathcal D$; group size $g=d_h=128$; nominal rank $k$.
\Ensure W4A4KV4 model $\widehat{\mathcal M}$
        and structured rotation factors
        $\{W_s,Y_s,\Pi_s,D_s\}_{s\in\mathcal J}$.


\PQStage{PQTOurs}{PQTInk}
{I. Spectral alignment: Householder / compact-WY}

\State
$\mathcal J
 \gets
 \{1\}\cup
 \{(2,\ell),(4,\ell):\ell=1,\ldots,L\}$
\PQNote{residual, value, down-input sites}

\State
$\{\Sigma_s\}_{s\in\mathcal J}
 \gets
 \Call{CollectMoments}{\mathcal M,\mathcal D},
 \qquad
 \Sigma_s=N_s^{-1}X_s^\top X_s$
\PQNote{uncentered}

\State
$k_1\gets\min(k,d/g),
 \quad
 k_{2,\ell}\gets1,
 \quad
 k_{4,\ell}\gets\min(k,d_{\mathrm{ff},\ell}/g)$
\PQNote{per-site rank $k_s$, capped by the slot count $d_s/g$}

\ForAll{$s\in\mathcal J$}

    \State
    $[v_{s,1},\ldots,v_{s,k_s}]
     \gets\PQOp{TopEig}_{k_s}(\Sigma_s),
     \qquad
     W_s,Y_s\gets[\,],[\,]$

    \For{$i=1,\ldots,k_s$}
    \PQNote{anchor $t_i=1+(i-1)g$: first coordinate of group $i$}

        \State
        $\delta_i
         \gets
         v_{s,i}-W_s(Y_s^\top v_{s,i})
         -e_{t_i}$

        \If{$\|\delta_i\|_2\ge10^{-7}$}
        \PQNote{skip if already anchored}

            \State
            $h_i\gets\delta_i/\|\delta_i\|_2$

            \State
            $W_s
             \gets
             [\,W_s-2h_i(h_i^\top W_s),\;2h_i\,],
             \quad
             Y_s\gets[\,Y_s,\;h_i\,]$

        \EndIf

    \EndFor

    \State
    $G_s\equiv I-W_sY_s^\top,
     \qquad
     \epsilon_s
     \gets
     \PQOp{diag}(G_s\Sigma_sG_s^\top)$
    \PQNote{per-coordinate energy after $G_s$}

    \State
    $\Pi_s
     \gets
     \Call{AnchorBalance}{\epsilon_s,k_s,g},
     \quad
     D_s\gets\PQOp{diag}(\xi_s),
     \quad
     \xi_s\in\{-1,+1\}^{d_s}$
    \PQNote{anchors fixed; seeded signs}

    \State
    $\textcolor{PQTInk}{
       R_s\equiv H_gD_s\Pi_s(I-W_sY_s^\top)
    }$
    \PQNote{%
      $R_sv_{s,i}=\pm u_i$,
      $u_i=\mathbf1_{\mathcal I_i}/\sqrt g$%
    }

\EndFor


\PQStage{PQTOffQ}{PQTWarm}
{II. Fold weights and register $R_1$, $R_2$, $R_3$, $R_4$}

\State
$\mathcal M\gets\Call{FuseRMSNorm}{\mathcal M},
 \qquad
 R_{3,\ell}\gets I\quad(\forall\ell)$
\PQNote{no post-RoPE query/key rotation; keys quantized per channel}

\State
$E\gets R_1E,
 \qquad
 A_{\mathrm{lm}}\gets A_{\mathrm{lm}}R_1^\top$
\PQNote{embedding and head}

\State
$A_{\mathrm{read}}
 \gets A_{\mathrm{read}}R_1^\top,
 \qquad
 A_{\mathrm{write}}
 \gets R_1A_{\mathrm{write}}$
\PQNote{every linear reading from or writing to the residual}

\For{$\ell=1,\ldots,L$}

    \State
    $A_{\mathrm{o},h}^{(\ell)}
     \gets
     A_{\mathrm{o},h}^{(\ell)}R_{2,\ell}^\top
     \quad(\forall h),
     \qquad
     A_{\mathrm{down}}^{(\ell)}
     \gets
     A_{\mathrm{down}}^{(\ell)}R_{4,\ell}^\top$
    \PQNote{$R_2$ into the output projection, $R_4^\top$ into the down projection}

    \State
    $T_\ell\gets D_{4,\ell}\Pi_{4,\ell},
     \quad
     A_{\mathrm{gate}}^{(\ell)}
     \gets
     \Pi_{4,\ell}A_{\mathrm{gate}}^{(\ell)},
     \quad
     A_{\mathrm{up}}^{(\ell)}
     \gets
     T_\ell A_{\mathrm{up}}^{(\ell)}$
    \PQNote{signed permutation folded; signs only on the up branch}

    \State
    $\widetilde W_\ell\gets T_\ell W_{4,\ell},
     \qquad
     \widetilde Y_\ell\gets T_\ell Y_{4,\ell}$
    \PQNote{$\widetilde G_\ell=T_\ell G_{4,\ell}T_\ell^\top$}

\EndFor

\State
$\mathsf V_\ell(v)\equiv R_{2,\ell}v$
\PQNote{applied after the value projection, shared across KV heads}

\State
$\textcolor{PQTInk}{
   \mathsf D_\ell(h_T)
   \equiv
   H_g\!\left[
     h_T-\widetilde W_\ell
     (\widetilde Y_\ell^\top h_T)
   \right]
}$
\PQNote{online $R_4$: rank-$k_s$ correction and block Hadamard}

\State
$\mathcal M_{\mathrm{rot}}
 \gets
 \Call{AttachRotations}
      {\mathcal M,\{\mathsf V_\ell,\mathsf D_\ell\}_\ell}$


\PQStage{PQTRef}{PQTLilac}
{III. Quantize weights offline; activations and KV online}

\State
$\widehat{\mathcal M}
 \gets
 \PQOp{GPTQ}_{4,g,\mathrm{asym}}
 (\mathcal M_{\mathrm{rot}};\mathcal D)$
\PQNote{A/KV quantization off during GPTQ}

\State
$\mathsf A(y)\equiv\mathcal Q_{4,g}(y)$
\PQNote{each quantized linear input}

\State
$\mathsf{KV}:\quad
 \widehat K_{\mathrm{old}}
 \gets
 \mathcal Q_{4,32}^{\mathrm{time}}(K_{\mathrm{old}}),
 \qquad
 \widehat V'_{\mathrm{old}}
 \gets
 \mathcal Q_{4,g}^{\mathrm{head}}(V'_{\mathrm{old}})$
\PQNote{keys per channel over 32 tokens; values per token over the head}

\State
$K_{\mathrm{current\ chunk}},
 \ V'_{\mathrm{last}\ 32}:
 \quad\text{full precision}$

\State \Return
$\Call{AttachQuantizers}
      {\widehat{\mathcal M},\mathsf A,\mathsf{KV}}$


\PQStage{PQTMint}{PQTGreen}
{IV. Grouped asymmetric INT4 with fp16 scale and offset}

\Function{$\mathcal Q_{4,g}$}{$y$}

    \ForAll{token/group vectors $y^{(j)}\in\mathbb R^g$}

        \State
        $z_j
         \gets
         \PQOp{fp16}(\min y^{(j)}),
         \qquad
         s_j
         \gets
         \PQOp{fp16}\!\left(
           \frac{\max y^{(j)}-\min y^{(j)}}{15}
         \right)$
        \PQNote{a group-constant level enters $z_j$, not $s_j$}

        \State
        $s_j\gets1\quad\text{if }s_j\le0$
        \PQNote{positive fallback scale}

        \State
        $\textcolor{PQTGreen}{
          q^{(j)}
          \gets
          \PQOp{clip}_{[0,15]}\!\left(
            \PQOp{round}\!\left(
              \frac{y^{(j)}-z_j\mathbf1_g}{s_j}
            \right)
          \right)
        }$

        \State
        $\textcolor{PQTGreen}{
          \widehat y^{(j)}
          \gets
          s_jq^{(j)}+z_j\mathbf1_g
        }$

    \EndFor

    \State \Return $\widehat y$
    \PQNote{QDQ form; codes and metadata are $(q,s,z)$}

\EndFunction

\end{algorithmic}
\endgroup
\end{algorithm}


\section{Theoretical Foundations, Proofs, and Additional Method Details}
\label{app:theoretical_foundations}
\label{app:method_details}

This appendix collects the proofs and implementation details supporting
\Cref{sec:method}; the range law, its metadata accounting, and its
empirical diagnostics are in \Cref{app:range_law}.
Exact statements assume real arithmetic unless finite precision is
explicitly discussed.
Orthogonal maps act on column vectors; row-batched activations are
transformed by right multiplication with the transpose.

\subsection{Group-Constant Geometry and Metadata Precision}
\label{app:group_constant_geometry}
\label{app:method_affine}

For a group $y\in\mathbb{R}^g$, let
$\Delta(y)=\max_i y_i-\min_i y_i$. For every scalar $c$,
\begin{equation}
    \min_i(y_i+c)=\min_i y_i+c,
    \qquad \Delta(y+c\mathbf{1}_g)=\Delta(y).
\end{equation}
Consequently, for the ideal affine quantizer with
$z(y)=\min_i y_i$ and $s(y)=\Delta(y)/15>0$,
\begin{equation}
    q(y+c\mathbf{1}_g)=q(y),
    \qquad
    \hat y(y+c\mathbf{1}_g)=\hat y(y)+c\mathbf{1}_g.
\end{equation}
A constant group is represented by its offset, with $s=1$ and $q=0$.

The embedded group indicators have disjoint supports and unit norm, so
$U^\top U=I_M$. The restriction of $UU^\top y$ to group $j$ is its arithmetic
mean repeated $g$ times. Subtracting this component is a group-common shift,
which proves \Cref{eq:range_residual_exact}. The projection is an analytical
decomposition, not an additional mean-subtraction operation in the algorithm.
In particular, the stored min-based offset need not equal the group mean:
if $y^{(j)}=c_j\mathbf{1}_g+r^{(j)}$, then
$z_j=c_j+\min_i r_i^{(j)}$.

The implementation uses $\tilde s=\operatorname{fp16}(s)$ and
$\tilde z=\operatorname{fp16}(z)$ before computing codes. In general,
$\operatorname{fp16}(z+c)\neq\operatorname{fp16}(z)+c$, so exact code invariance
does not extend to arbitrary shifts with finite-precision metadata.
The underlying range identity remains valid in real arithmetic; metadata
rounding is a separate numerical error source. Zero or underflowed scales
are handled by the implementation's nonzero-scale guard. These qualifications
are why the main text does not describe the offsets as an unlimited,
exactly lossless storage channel.

\paragraph{A four-feature illustration.}
Mapping $v_1=(1,-1,1,-1)^\top/2$ to
$u_1=(1,1,1,1)^\top/2$ sends the dominant component
$(10,-10,10,-10)^\top$ to $(10,10,10,10)^\top$.
Its energy is unchanged, but its contribution to range falls from 20
to zero. With transformed residual $(-0.20,-0.03,0.03,0.20)^\top$, the
quantizer receives $(9.80,9.97,10.03,10.20)^\top$, with ideal offset
$9.80$ and step $0.40/15$. The gain comes from rotating a varying
pattern into a shared level, not translating an unchanged group.
The group mean is $10$, illustrating why it differs from the stored
min-based offset.

\paragraph{Residual energy bounds quantization range.}
Let $e=(I-P_{\mathcal S})y$. For any nonconstant group $e^{(j)}$,
choose distinct indices $p$ and $q$ attaining its maximum and minimum.
Then
\begin{equation}
    \operatorname{range}(e^{(j)})^2
    =(e^{(j)}_p-e^{(j)}_q)^2
    \leq 2\big((e^{(j)}_p)^2+(e^{(j)}_q)^2\big)
    \leq 2\|e^{(j)}\|_2^2.
\end{equation}
The constant case is immediate. Summing over groups and using
\Cref{eq:range_residual_exact} proves \Cref{eq:range_energy_bound}.
The same argument applies to $e_k=(I-U_kU_k^\top)y$, because the removed
component is group-constant even when $k<M$. Thus maximizing
$\mathcal J_k$ minimizes the corresponding residual-energy upper bound,
not necessarily the realized range.

Under ideal min--max quantization, nearest-grid rounding incurs coordinate
error at most $s_j/2$. Consequently,
\begin{equation}
    \|\hat y-y\|_2^2
    \leq \frac{g}{4\cdot15^2}
         \sum_{j=1}^{M}\operatorname{range}(y^{(j)})^2
    \leq \frac{g}{2\cdot15^2}\|(I-P_{\mathcal S})y\|_2^2.
\end{equation}
These are worst-case bounds in ideal arithmetic, not equality statements
or population predictions. Finite-precision metadata and subsequent weight
quantization require separate treatment.

\paragraph{Offset precision.}
\label{app:metadata_precision}
Range-nullity is exact under ideal affine quantization; the stored fp16
offset carries a relative rounding error of at most $2^{-11}$, which grows
with the aligned level.
Table~\ref{tab:app_offset_precision} measures this error at fixed codes:
for every group of the activation-only runs of Appendix~\ref{app:offset},
the offset is re-stored in fp32 or in bf16 (seven fraction bits) while
codes and scales are held bit-identical, and the rounding error of the
fp16 offset is compared with the quantization step of its group.
Aligned levels reach ten to twenty times the within-group range at the
first layers, yet the rounding error has a median of $0.001$ steps, a
99th percentile of $0.025$ steps, and exceeds half a step in fewer than
one group in a million; the only such groups are BOS positions at the
first down-projection input.
Relative to the step that the same level would cost if it entered the
range unaligned, $2|c|/15$, the rounding error is about one percent.
Perplexity is insensitive to the offset's precision: fp32 and even bf16
offsets change it by less than $0.001$, for \pq{} and Hadamard alike.
The level is therefore not stored losslessly, but its storage error is
two orders of magnitude below the alternative and invisible end to end.

\begin{table}[t]
\centering
\captionsetup{font=small,skip=4pt}
\caption{\textbf{fp16 offset rounding at fixed codes.}
\pq{} at $g=128$, $k=\max$, activation-only, three seeds.
Offset error is $|z_j-\mathrm{fp16}(z_j)|$ in units of the group's
step; the unaligned cost is $2|c|/15$ for the same level $c$.
$\Delta$PPL rows re-store only the offset at the stated precision.}
\label{tab:app_offset_precision}
\begingroup
\fontsize{7pt}{8pt}\selectfont
\setlength{\tabcolsep}{1.8pt}
\renewcommand{\arraystretch}{1.00}
\setlength{\aboverulesep}{1.8pt}
\setlength{\belowrulesep}{1.8pt}
\arrayrulecolor{black}
\begin{tabularx}{\linewidth}{
    >{\raggedright\arraybackslash}p{0.44\linewidth}
    *{2}{>{\raggedleft\arraybackslash}X}
}
\toprule
\rowcolor{PQTBand}
& \textbf{Qwen3-4B-Base} & \textbf{Llama-3.2-3B} \\
\midrule
$|z|/\mathrm{range}$, median / P99 / max (worst layer)
& 0.51 / 10.0 / 21.5 & 0.52 / 5.8 / 10.2 \\
Offset error / step, median / P99 (worst layer)
& 0.0012 / 0.025 & 0.0012 / 0.017 \\
Groups with error $>0.5$ step, overall
& $8\times10^{-5}\%$ & $5\times10^{-5}\%$ \\
\quad worst layer/site
& $0.003\%$ & $0.0008\%$ \\
Median (offset error)/(unaligned cost)
& 0.0085 & 0.0093 \\
\pqtdashline
$\Delta$PPL, fp32 offset
& $+0.0008$ & $+0.0007$ \\
$\Delta$PPL, bf16 offset
& $+0.0007$ & $+0.0007$ \\
\bottomrule
\end{tabularx}
\endgroup
\end{table}

\subsection{Proof of Optimal Spectral Alignment}
\label{app:spectral_optimality}
\label{app:method_spectral}

Let $B=R^\top U_k$. Orthogonality of $R$ implies $B^\top B=I_k$, and
\begin{equation}
    \mathcal J_k(R)
    =\operatorname{Tr}(B^\top\Sigma B)
    =\sum_{i=1}^d\lambda_i\gamma_i,
    \qquad \gamma_i=\|B^\top v_i\|_2^2.
\end{equation}
Because $BB^\top$ is a rank-$k$ orthogonal projector,
$0\leq\gamma_i\leq1$ and $\sum_i\gamma_i=k$. Since the eigenvalues are
nonincreasing, allocating these $k$ units of mass to the largest eigenvalues
gives
\begin{equation}
    \mathcal J_k(R)\leq\sum_{i=1}^k\lambda_i.
\end{equation}
Equality holds when $\operatorname{span}(B)=\operatorname{span}(V_k)$,
which is achieved by an orthogonal map carrying $V_k$ onto $U_k$.
This proves \Cref{prop:optimal_alignment}, a direct application of the
Ky Fan maximum principle \citep{fan1949theorem,fan1950theorem}.
Eigenvalue ties can yield multiple maximizers; no uniqueness is asserted.

For $k=M$, energy preservation gives
\begin{equation}
    \min_{R^\top R=I}
    \mathbb{E}\|(I-P_{\mathcal S})Rx\|_2^2
    =\operatorname{Tr}(\Sigma)-\sum_{i=1}^M\lambda_i.
\end{equation}
For $k<M$, optimality applies to the selected target $U_k$, not the
unrestricted $M$-dimensional target. If calibration returns an orthonormal
approximation $\widehat V_k$, exact alignment of that approximation captures
$\operatorname{Tr}(\widehat V_k^\top\Sigma\widehat V_k)$ in the selected target.
The discrepancy from the spectral optimum is due to eigenspace estimation,
not the Householder representation. Likewise, optimizing the empirical
calibration moment is distinct from guaranteeing the population optimum.

\subsection{Householder Construction and Compact Representation}
\label{app:householder_construction}
\label{app:method_construction}

Let $t_i=1+(i-1)g$ and $e_{t_1},\ldots,e_{t_k}$ be distinct
coordinate anchors; set $G_0=I$.
At step $i$, write $b_i=G_{i-1}v_i$. If $b_i=e_{t_i}$, skip the reflection.
Otherwise, define
\begin{equation}
    h_i=\frac{b_i-e_{t_i}}{\|b_i-e_{t_i}\|_2},
    \qquad \mathcal H_i=I-2h_i h_i^\top,
    \qquad G_i=\mathcal H_iG_{i-1}.
\end{equation}
The vectors $b_i$ and $e_{t_i}$ have unit norm, giving
$\mathcal H_i b_i=e_{t_i}$. For $\ell<i$, orthogonality implies
$b_i^\top e_{t_\ell}=v_i^\top v_\ell=0$. Distinct anchors also satisfy
$e_{t_i}^\top e_{t_\ell}=0$. Hence $h_i$ is orthogonal to every previously
aligned anchor, and $\mathcal H_i$ preserves them. Induction yields
$G_kv_i=e_{t_i}$ for every $i\leq k$.

Choose $\Pi e_{t_i}=e_{p_i}$, where $p_i$ is the first coordinate of group
$j_i$. A diagonal sign matrix maps this anchor to $\pm e_{p_i}$, and the
normalized block Hadamard maps it to $\pm u_{j_i}$. Therefore
$H_gD\Pi G_kv_i=\pm u_{j_i}$, proving that the structured construction attains
the selected-subspace optimum for exact eigenvectors. Balancing the other
coordinates does not change these anchor constraints.

To obtain compact factors, suppose $G_{i-1}=I-W_{i-1}Y_{i-1}^\top$. Then
\begin{equation}
    G_i=I-2h_i h_i^\top-
        \mathcal H_iW_{i-1}Y_{i-1}^\top
        =I-W_iY_i^\top,
\end{equation}
where
\begin{equation}
    W_i=[\mathcal H_iW_{i-1},\,2h_i],
    \qquad Y_i=[Y_{i-1},\,h_i].
\end{equation}
Starting with empty factors proves a representation with at most $k$ columns,
consistent with compact Householder representations
\citep{schreiber1989storage}. Thus $Gx=x-W(Y^\top x)$ and, for row batches,
$XG^\top=X-(XY)W^\top$. The orthogonal matrix $G$ is full rank; it is the
correction $I-G$ that has rank at most $k$.

\paragraph{Coordinate placement and finite-precision checks.}
The selected anchors occupy the first coordinates of the first $k$ groups.
When $k<M$, unused first-coordinate slots receive the lowest-energy
available coordinates, using energies estimated after $G$ from the
calibration data. Remaining coordinates are sorted by decreasing energy
and assigned greedily to the group with the least accumulated residual
energy, subject to its $g-1$ non-anchor slots. Ties are deterministic.
This preserves the selected anchor constraints and avoids unintentionally
allocating high-energy coordinates to unused constant slots in reduced-rank
comparisons. It is a residual-balancing rule, not a separate guarantee of
optimal group range.

The implementation skips a reflection when
$\|b_i-e_{t_i}\|_2<10^{-7}$. The induction above establishes exact
alignment for orthonormal inputs in real arithmetic; the threshold,
estimated eigenspace, and floating-point products introduce separately
recorded numerical residuals. Signs may change $u_{j_i}$ to $-u_{j_i}$
without changing either its span or captured energy.

\subsection{Calibration and Numerical Approximation}
\label{app:method_calibration}

Wide activation sites use a randomized sketch with oversampling 16 rather
than an explicitly formed $d\times d$ second moment. Two second-moment
applications with orthogonalization are followed by a third pass for
Rayleigh--Ritz extraction and residual checks. Permutation-energy collection
is an additional calibration step, not part of the three eigensolver passes.
For sketch width $\ell$, streamed products cost
$\mathcal O(N_{\rm cal}d\ell)$ per pass, in addition to orthogonalization
and the smaller projected eigendecomposition. Head-dimensional value
moments are small enough to form explicitly and diagonalize directly.

Calibration is site-specific. The $R_1$ moment pools the outputs of
attention-input RMSNorms across layers, with equal token weighting.
Each $R_4$ uses that layer's gated down-projection inputs. Each value
moment pools tokens and KV heads within its layer; the resulting $R_2$
is shared across those heads. A global pooled $R_1$ objective must not be
confused with a separate per-layer optimum inferred from activation
diagnostics.

All factors and seeded signs are fixed before evaluation, without
gradient-based training. Ritz residuals assess eigenspace estimation,
while anchor and rotation checks assess its realization. Calibration
settings are fixed by the experimental protocol; rank controls alignment
capacity and online cost. These checks do not convert the empirical
calibration optimum into a guaranteed optimum for unseen activations.

\subsection{Gaussian Range Model and the Two-Factor Predictor}
\label{app:range_model}

Consider a conditional model for a group after mixing,
$y=c\mathbf{1}_g+\sigma Z$, where the coordinates of $Z$ are independent
standard normal variables. The common component does not affect range, so
\begin{equation}
    \mathbb{E}[\Delta(y)\mid c,\sigma]=\sigma\eta_g,
    \qquad
    \eta_g=2\mathbb{E}\max_{i\leq g}Z_i.
\end{equation}
Symmetry gives the second identity. If $\Phi$ and $\phi$ denote the standard
normal distribution and density, the maximum has distribution $\Phi(t)^g$,
and therefore
\begin{equation}
    \eta_g=2g\int_{-\infty}^{\infty}
        t\phi(t)\Phi(t)^{g-1}\,dt.
\end{equation}
The ideal mean quantization step is $\sigma\eta_g/15$. Assuming comparable
coordinate fluctuation scale across group sizes gives
\Cref{eq:group_size_factor}; the factor $\eta_g$ is determined by $g$,
without fitting a regression coefficient.

\section{Range Law and Metadata Budget}
\label{app:range_law}
\label{app:method_range}
\label{sec:theory}

\subsection{Range Prediction}
\label{app:range_prediction}

\paragraph{From residual energy to range.}
The alignment objective controls residual energy, whereas the quantizer
responds to within-group range.
For a group $v\in\mathbb{R}^g$, define
$e=v-\bar v\mathbf{1}_g$, where
$\bar v=g^{-1}\mathbf{1}_g^\top v$.
Subtracting a shared level leaves the range unchanged.
If $p$ and $q$ index a maximum and a minimum, respectively, then
\begin{equation}
    \operatorname{range}(v)^2
    =(e_p-e_q)^2
    \leq 2(e_p^2+e_q^2)
    \leq 2\|e\|_2^2,
    \label{eq:app_group_range_bound}
\end{equation}
with the constant-group case immediate.
Summing over groups yields
\begin{equation}
    \sum_{j=1}^{M}\operatorname{range}(y^{(j)})^2
    \leq 2\|(I-P_{\mathcal S})y\|_2^2.
    \label{eq:range_energy_bound}
\end{equation}
This deterministic inequality requires no distributional assumption.
It motivates residual energy as a surrogate for controlling range:
reducing the right-hand side tightens the bound but does not guarantee
a decrease in every realized group range.
Predicting the typical range additionally requires a model of the
residual distribution.

\paragraph{Energy factor.}
Let $\mathcal E=\operatorname{Tr}(\Sigma)>0$ and
$f_k=\sum_{i\leq k}\lambda_i/\mathcal E$ be the deliberately aligned
energy fraction in the ideal construction.
The decomposition in \Cref{eq:eigenspace_decomposition} and orthogonality
of $R$ give
\begin{equation}
    \mathbb{E}\|Rr\|_2^2
    =\mathbb{E}\|r\|_2^2
    =(1-f_k)\mathcal E.
    \label{eq:app_residual_energy}
\end{equation}
Since $RV_ka=U_ka\in\mathcal S$,
$(I-P_{\mathcal S})Rx=(I-P_{\mathcal S})Rr$, and hence
\begin{equation}
    \mathbb{E}\sum_{j=1}^{M}
        \operatorname{range}((Rx)^{(j)})^2
    \leq 2(1-f_k)\mathcal E.
    \label{eq:app_expected_range_bound}
\end{equation}
At $k=M$, $Rr$ lies entirely outside $\mathcal S$.
At $k<M$, unused constant directions may absorb additional residual
energy, so $f_k$ need not equal the total fraction captured by
$\mathcal S$.
For an estimated eigenspace, the corresponding captured fraction is
$\operatorname{Tr}(\widehat V_k^\top\Sigma\widehat V_k)/\mathcal E$;
the eigenvalue sum is exact for the moment whose leading
eigenvectors are used.

Under the approximation that mixed residuals retain comparable
normalized shapes and their scales change proportionally across
matched token--group observations, their amplitude, and hence range,
decreases as $\sqrt{1-f_k}$.
Aggregate energy alone does not imply this scaling of mean range.
Hadamard mixing motivates the approximation but does not guarantee
it for arbitrary activations.

\paragraph{What the inequality does bound.}
Define the mean ideal INT4 step of a rotation $R$ as
$\bar s_R=\frac{1}{15M}\mathbb E_x\sum_{j}\operatorname{range}((Rx)^{(j)})$.
Cauchy--Schwarz over groups and Jensen's inequality applied to
\Cref{eq:app_expected_range_bound} give the distribution-free bound
\begin{equation}
    \bar s_R\le\frac{1}{15}\sqrt{\frac{2(1-f_k)\mathcal E}{M}}.
    \label{eq:app_mean_step_bound}
\end{equation}
Its prefactor is set by the total energy, not by the measured Hadamard
step; replacing it with $s_{\rm H}(g)$ is the modeling step of the
predictor below, not a consequence of the inequality.

\paragraph{Exact decomposition.}
Let $\sigma_R^2=\mathbb E\|(I-P_{\mathcal S})Rx\|_2^2/d$ be the residual
energy per coordinate under $R$, $f_R$ the fraction of energy in
$\mathcal S$ under $R$, and $\kappa_R=15\,\bar s_R/\sigma_R$ the mean
group range in units of the residual r.m.s., a crest factor.
By definition,
\begin{equation}
    \frac{\bar s_{\rm PQ}}{\bar s_{\rm H}}
    =\frac{\kappa_{\rm PQ}}{\kappa_{\rm H}}
     \sqrt{\frac{1-f_{\rm PQ}}{1-f_{\rm H}}},
    \label{eq:app_crest_identity}
\end{equation}
with no assumption.
At full capacity and exact alignment $f_{\rm PQ}=f_k$, and an isotropic
Hadamard captures $f_{\rm H}\approx1/g$.
The predictor is the special case $\kappa_{\rm PQ}=\kappa_{\rm H}$,
$f_{\rm H}=0$: it assumes the residual keeps its crest factor.
\Cref{eq:app_crest_identity} therefore locates every departure of the
predictor in one measurable ratio, reported in
\Cref{app:rangelaw_persite}.

\paragraph{Group-size factor.}
Let $Z_i\overset{\rm iid}{\sim}\mathcal N(0,1)$ and define
\begin{equation}
    \eta_g
    =\mathbb{E}\!\left[
        \max_{i\leq g}Z_i-\min_{i\leq g}Z_i
    \right].
    \label{eq:app_gaussian_range_definition}
\end{equation}
A centered Gaussian surrogate
$e=\sigma(Z-\bar Z\mathbf{1}_g)$ satisfies
$\mathbb{E}\operatorname{range}(e)=\sigma\eta_g$, because subtracting
the sample mean leaves the range unchanged.
This formulation respects the zero-sum constraint on group residuals;
the centered coordinates themselves are not independent, and
$\mathbb{E}\|e\|_2^2=(g-1)\sigma^2$.

Writing $\phi$ and $\Phi$ for the standard normal density and
distribution function, the maximum has distribution $\Phi(t)^g$.
Symmetry and the order-statistic density give
\begin{align}
    \eta_g
    &=2g\int_{-\infty}^{\infty}
       t\,\phi(t)\Phi(t)^{g-1}\,dt \nonumber\\
    &=2\int_0^\infty
       \left[1-\Phi(t)^g-(1-\Phi(t))^g\right]dt.
    \label{eq:app_gaussian_range_integral}
\end{align}
We compute the finite-$g$ reference by numerical quadrature:
$\eta_{64}\approx4.687467$,
$\eta_{128}\approx5.189195$, and
$\eta_{256}\approx5.653727$.
Although $\eta_g\sim2\sqrt{2\ln g}$ asymptotically
\citep{david2004order}, the reported comparisons use these finite-$g$
values rather than the leading asymptotic expression.

Under Gaussian mixing with comparable scale parameters at two group
sizes, the mean ideal steps of the paired Hadamard references satisfy
\begin{equation}
    \frac{s_{\rm H}(g_2)}{s_{\rm H}(g_1)}
    \approx\frac{\eta_{g_2}}{\eta_{g_1}}.
    \label{eq:group_size_factor}
\end{equation}
Here an ideal INT4 step is the group range divided by $15$.
If the Gaussian scale parameters differ, their ratio also enters
\Cref{eq:group_size_factor}; comparability of these scales is an
assumption rather than a consequence of the order-statistic formula.

\paragraph{Predictor.}
Under the shape assumption above, and neglecting incidental capture by
the reference and unused target slots, we use the baseline-relative
predictor
\begin{equation}
    \widehat s(g,k)=s_{\rm H}(g)\sqrt{1-f_k}.
    \label{eq:range_predictor}
\end{equation}
The measured reference $s_{\rm H}(g)$ absorbs both the group-size
factor and the INT4 denominator $15$.
It is evaluated at the same reference site and group size using
matched evaluation tokens, while $f_k$ is obtained from the calibration
spectrum.
No multiplicative constant is fitted.
Unlike \Cref{eq:app_expected_range_bound,eq:app_mean_step_bound},
\Cref{eq:range_predictor} is not a distribution-free upper bound on the
mean step: it is the special case of \Cref{eq:app_crest_identity} with
$\kappa_{\rm PQ}=\kappa_{\rm H}$ and $f_{\rm H}=0$.
Unused target slots, nonuniform token or group scales, correlations,
heavy tails, and calibration mismatch all enter through that crest-factor
ratio, and the residual-shape approximation does not follow from spectral
optimality.
The pooled regression and held-out MoE expert visualization in
\Cref{app:method_range_diagnostics} assess the empirical trend separately;
they neither prove nor parameterize \Cref{eq:range_predictor}.
The per-site check there measures the crest-factor ratio directly: it is
within $3\%$ of one at the q/k/v input and $0.87$--$0.92$ at the
down-projection input.

\subsection{Metadata Accounting and Extended-Affine Controls}
\label{app:method_metadata}

\paragraph{Two resources, one budget.}
Finer grouping increases both the number of local scales and the
dimension of $\mathcal S$: halving $g$ doubles both counts.
Scale refinement changes local quantization resolution, while additional
constant directions help only to the extent that they capture relevant
energy.
With $g$ INT4 codes, one FP16 scale, and one FP16 offset per group,
the logical storage is
\begin{equation}
    b_{\rm eff}(g,1)
    =\frac{4g+16+16}{g}
    =4+\frac{32}{g}
    \quad\text{bits per value}.
    \label{eq:app_standard_affine_budget}
\end{equation}
This gives $4.5$ at $g=64$, $4.25$ at $g=128$, and $4.125$ at $g=256$.
Raising $k$ at fixed $g$ leaves this per-value format budget unchanged
and instead increases factor storage and, at online sites, transform work.
These counts cover codes and per-group metadata.
Shared transform factors, packing or allocation overhead, temporary
buffers, and full-precision cache tails are separate deployment costs.

\paragraph{Metadata precision.}
The range predictor describes ideal affine arithmetic.
For a nonconstant group, write
$z=\min_i v_i$, $s=\operatorname{range}(v)/15$, and denote the
stored FP16 parameters by
$\widetilde z=z+\delta_z$ and $\widetilde s=s+\delta_s$.
The encoder uses these rounded parameters, so its integer codes
may differ from those obtained with the ideal grid.
For finite $\widetilde z$ and positive finite $\widetilde s$,
with the remaining arithmetic evaluated exactly, nearest-point
encoding on the rounded grid satisfies
\begin{equation}
    |\hat v_i-v_i|
    \leq \frac{s}{2}+|\delta_z|+15|\delta_s|.
    \label{eq:app_metadata_rounding_bound}
\end{equation}
To see this, each ideal grid point $z+qs$ moves by
$\delta_z+q\delta_s$, whose magnitude is at most
$|\delta_z|+15|\delta_s|$ for $q\in\{0,\ldots,15\}$.
The nearest point on the rounded grid is no farther than the perturbed
ideal nearest point.
Additional floating-point arithmetic errors are outside this bound.
Thus a shared level remains exactly range-neutral, although a large
offset can increase absolute FP16 rounding error.
Constant groups have zero ideal range and use a positive fallback
scale for encoding; metadata precision still limits reconstruction.

\paragraph{Extended-affine controls.}
To distinguish representation capacity from scale resolution, the
controls retain the offset and add $m-1$ FP16 coefficients along fixed,
orthonormal, nonconstant Walsh directions.
Let $B\in\mathbb{R}^{g\times(m-1)}$ satisfy
$B^\top B=I$ and $B^\top\mathbf{1}_g=0$.
Representing a component $Bc$ separately leaves a remainder to be
encoded on the affine INT4 grid, with reconstruction
\begin{equation}
    \hat v=\widetilde s\,q+\widetilde z\,\mathbf{1}_g
           +B\widetilde c,
    \qquad q\in\{0,\ldots,15\}^g.
    \label{eq:app_extended_affine_reconstruction}
\end{equation}
The explicitly represented subspace
$\operatorname{span}(\mathbf{1}_g,B)$ has dimension $m$.
The additional directions generally have nonzero range:
they are represented through extra coefficients, rather than becoming
range-null directions of the original affine quantizer.
The offset sets the origin of the grid for the remainder and need
not equal the original group mean.

The fixed Walsh basis is shared, so there is no per-group basis cost.
Counting the codes, scale, offset, and additional coefficients gives
\begin{equation}
    b_{\rm eff}(g,m)
    =\frac{4g+16+16+16(m-1)}{g}
    =4+\frac{16(m+1)}{g}.
    \label{eq:extended_affine_budget}
\end{equation}
Coefficient precision also matters:
if $\delta c=\widetilde c-c$, then
$\|B\delta c\|_2=\|\delta c\|_2$.
More directions can increase captured energy without guaranteeing
a smaller quantization error or better perplexity at fixed storage.

\paragraph{Measured comparisons.}
At the shared $4.25$-bit budget, $(g,m)=(128,1)$ outperforms $(256,3)$
in activation-only perplexity on both tested Llama checkpoints.
Moreover, \pq{} at $(256,1)$ outperforms Hadamard at $(128,1)$ while
using fewer logical activation bits
(\Cref{tab:app_ablations}, \Cref{fig:budget_deployment}).
Group size therefore determines both local resolution and the available
constant subspace, while rank controls how much leading energy is
explicitly assigned to that subspace.
The matched-budget comparison changes both $g$ and $m$; it evaluates
their trade-off rather than isolating either factor.
These observations do not establish a universal optimum over quantizer
formats, and neither larger rank nor more represented directions alone
guarantees better perplexity at a fixed resource budget.
In the same study, the step ordering predicted by \Cref{eq:range_predictor}
agrees with the mean-PPL ordering over the six bit-accounted \pq{}
configurations on each Llama checkpoint (Spearman correlation $1.0$).
This is a ranking observation for those configurations, not a measure
of absolute prediction accuracy.

\begin{figure}[!t]
    \centering
    \captionsetup{font=small,skip=4pt}
    \includegraphics[width=\linewidth]{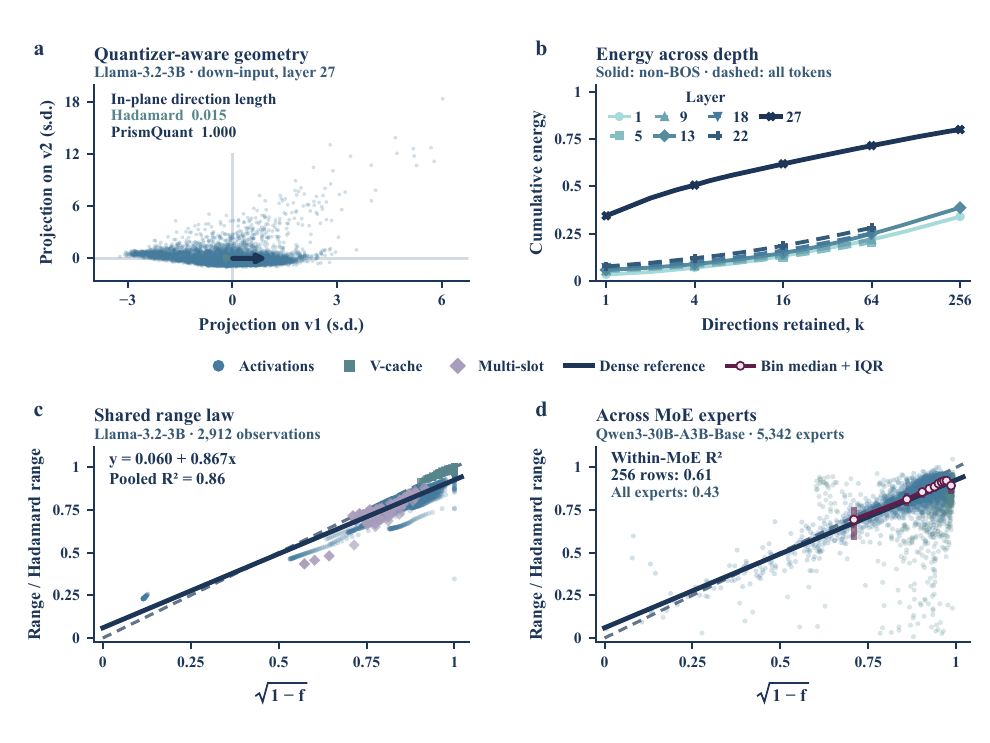}
    \caption{\textbf{From null-space geometry to a shared range law.}
    \textbf{(a)}~Standardized token projections at the Llama-3.2-3B layer-27 down input; arrows are the group-constant direction pulled back through each rotation.
    \textbf{(b)}~Cumulative energy in the leading directions at seven layers.
    \textbf{(c)}~Group range relative to paired Hadamard versus $\sqrt{1-f}$
    for 2,520 activation, 280 V-cache, and 112 multi-slot observations under one pooled fit; dashed is the identity.
    \textbf{(d)}~The same fitted line on 5,342 Qwen3-30B-A3B experts held out of the fit: bin medians (Q1--Q3 bars) follow the dense reference line, and the scatter is widest for experts with fewer than 2,048 tokens.}
    \label{fig:geometry_range}
\end{figure}

\subsection{Empirical Diagnostics of the Range Law}
\label{app:method_range_diagnostics}

\paragraph{Geometry and pooled range trend.}
\Cref{fig:geometry_range} collects the geometric, spectral, and range
visualizations.
Panel (a) shows standardized token projections and the pulled-back
group-constant directions, while panel (b) shows cumulative leading
energy at seven layers.
The pooled dense-reference diagnostic in panel (c) contains 2,520
activation observations, 280 V-cache observations, and 112
extended-affine controls, for 2,912 observations in total.
Let $\rho$ denote group range relative to its paired Hadamard reference.
We retain the source-defined absorbed-energy fractions $f$, which need
not coincide with the deliberately aligned spectral fraction $f_k$
for every configuration.
The displayed least-squares line is
\begin{equation}
    \rho=0.0598022+0.8665148\sqrt{1-f},
    \qquad R^2=0.861389.
    \label{eq:app_pooled_range_fit}
\end{equation}
This is one pooled descriptive fit, not a separate regression per family.
The dashed identity line shows $\rho=\sqrt{1-f}$ for comparison.
The fitted intercept and slope are not used in
\Cref{eq:range_predictor}, which has no fitted multiplicative coefficient.
Accordingly, the reported $R^2$ describes the regression fit and is
not an accuracy score for the unfitted predictor.

\paragraph{Held-out expert diagnostic.}
Panel (d) applies the unchanged dense-reference line to 5,342
Qwen3-30B-A3B experts held out of that fit.
The per-expert points and binned medians with Q1--Q3 intervals expose
variability beyond the average trend, including the wider scatter
among experts with fewer than 2,048 calibration tokens.
The quartile intervals describe within-bin spread, not confidence
intervals for the medians.
The holdout concerns the pooled regression and does not imply that
the experts were excluded from their own rotation calibration.
This is an empirical transfer diagnostic, rather than an exact
consequence of the Gaussian range model or a universal per-expert
prediction guarantee.

\paragraph{Group-size diagnostic.}
On the two Llama checkpoints in the activation-only metadata study,
measured Hadamard step ratios for $256/128$ and $128/64$ at both q/k/v
and down-input sites differ from the finite-$g$ Gaussian ratios by
less than $0.2\%$.
The reference ratios use \Cref{eq:app_gaussian_range_integral}.
This agreement concerns the group-size factor alone, not the absolute
prediction error of the complete energy-and-group-size law.

\begin{figure}[t]
    \centering
    \captionsetup{font=small,skip=4pt}
    \includegraphics[width=\linewidth]{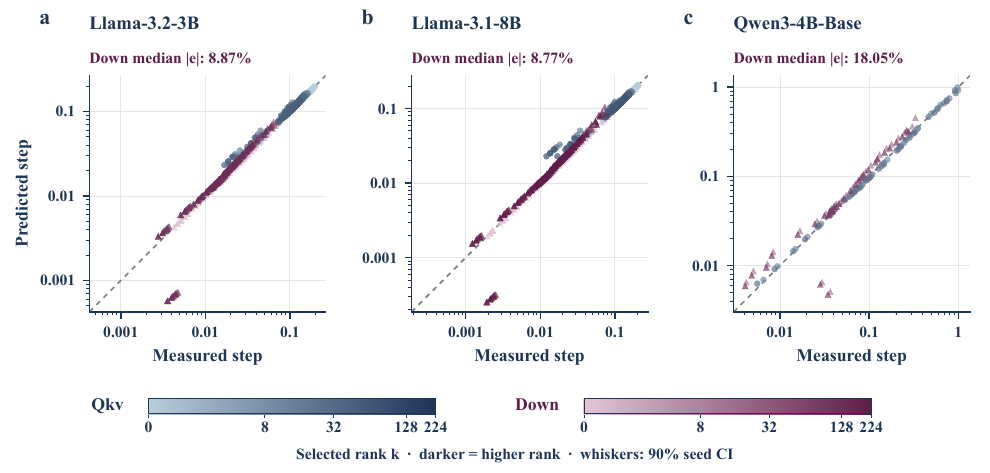}
    \caption{\textbf{Predicted versus measured quantization step.}
    Each point is one layer and site of the activation-only studies at
    $g=128$; whiskers are 90\% intervals over three paired seeds and the
    dashed line is the identity.
    Points above the line are over-predicted.
    The q/k/v input follows the identity closely at every rank; the
    down-projection input sits below it, more so on Qwen3-4B-Base; the
    ratio of measured to predicted step there is the crest-factor ratio
    of \Cref{eq:app_crest_identity}.}
    \label{fig:rangelaw_persite}
\end{figure}

\paragraph{Per-site prediction error.}
\label{app:rangelaw_persite}
The complete law is accurate at the q/k/v input and conservative at the
down-projection input.
Figure~\ref{fig:rangelaw_persite} and Table~\ref{tab:rangelaw_persite}
compare the predicted step $\widehat s(g,k)=s_{\rm H}(g)\sqrt{1-f_k}$
with the measured mean step at every layer, site, and rank of the
activation-only studies, with 90\% intervals over three paired seeds and
no fitted coefficient.
At $g=128$ and $k=\max$, the median absolute relative error is
$1$--$4\%$ at q/k/v but $9\%$ on both Llama models and $18\%$ on
Qwen3-4B-Base at the down-projection input, and its sign is systematic:
the measured step is smaller than predicted.
The bias has the direction the assumptions allow.
The law assumes that removing aligned energy leaves a residual of the
same shape, so that the range shrinks with the amplitude
$\sqrt{1-f_k}$; the leading directions are, however, the heavy-tailed
part of the activation, and removing them shrinks the extremes by more
than the r.m.s., most strongly on Qwen3-4B-Base, whose leading
directions are the most concentrated.
Incidental capture by the Hadamard reference would bias the prediction
the other way, so it is not the cause.
Two properties survive unchanged: the predicted ordering of the six
bit-accounted configurations matches the measured perplexity ordering
(\Cref{app:method_metadata}), and the group-size ratios agree to
$0.2\%$ (above).
In the terms of \Cref{eq:app_crest_identity}, the measured
$\kappa_{\rm PQ}/\kappa_{\rm H}$ is $0.98$--$1.03$ at q/k/v and
$0.87$--$0.92$ at the down-projection input: alignment removes the
heavy-tailed directions and leaves a flatter residual, the same
flattening that the anchor-column ablation of
Appendix~\ref{app:offset} identifies as the main source of the gain.
The predictor is therefore used to order configurations and as an
empirically conservative estimate at the down-projection input.
 
\begin{table}[t]
\centering
\captionsetup{font=small,skip=4pt}
\caption{\textbf{Per-site error of the range law at $g=128$.}
Relative error $e=(\widehat s-s_{\rm meas})/s_{\rm meas}$ of the
predicted step, activation-only protocol, three paired seeds; medians
and 90th percentiles of $|e|$ over layers, and the median signed error,
at the default operating point $k=8$ and at $k=\max$.
Positive signed error means the law over-predicts the step; the last
column is the crest-factor ratio of \Cref{eq:app_crest_identity},
$1/(1+\mathrm{median}\,e)$.}
\label{tab:rangelaw_persite}
\begingroup
\fontsize{7pt}{8pt}\selectfont
\setlength{\tabcolsep}{1.8pt}
\renewcommand{\arraystretch}{1.00}
\setlength{\aboverulesep}{1.8pt}
\setlength{\belowrulesep}{1.8pt}
\arrayrulecolor{black}
\begin{tabularx}{\linewidth}{
    >{\raggedright\arraybackslash}p{0.20\linewidth}
    >{\raggedright\arraybackslash}p{0.07\linewidth}
    *{5}{>{\raggedleft\arraybackslash}X}
}
\toprule
\rowcolor{PQTBand}
\textbf{Model} & \textbf{Site}
& \shortstack[r]{$k=8$\\median $|e|$}
& \shortstack[r]{$k=\max$\\median $|e|$}
& \shortstack[r]{$k=\max$\\P90 $|e|$}
& \shortstack[r]{$k=\max$\\median $e$}
& \shortstack[r]{$k=\max$\\$\kappa_{\rm PQ}/\kappa_{\rm H}$} \\
\midrule
Llama-3.2-3B & q/k/v & 1.7\% & 1.7\% & 8.9\% & \pqtworse{+1.7\%} & 0.98 \\
Llama-3.2-3B & down  & 11.1\% & 8.9\% & 15.0\% & \pqtworse{+8.4\%} & \pqtshare{0.92} \\
\pqtdashline
Llama-3.1-8B & q/k/v & 2.0\% & 1.3\% & 6.4\% & \pqtworse{+1.3\%} & 0.99 \\
Llama-3.1-8B & down  & 12.2\% & 8.8\% & 15.8\% & \pqtworse{+8.6\%} & \pqtshare{0.92} \\
\pqtdashline
Qwen3-4B-Base & q/k/v & 2.5\% & 3.5\% & 6.2\% & \pqtgain{-3.2\%} & 1.03 \\
Qwen3-4B-Base & down  & 21.2\% & 18.1\% & 52.2\% & \pqtworse{+15.4\%} & \pqtshare{0.87} \\
\bottomrule
\end{tabularx}
\endgroup
\end{table}
 
\section{Efficient Deployment and End-to-End Performance}
\label{app:prismquant_deployment}
\label{app:metadata_deployment} 

We evaluate whether the additional structure of \pq{} can be implemented
without sacrificing the practical benefits of low-bit inference.
Our implementation combines a compact low-rank rotation, Tensor Core
execution, pre-bound kernel launches, and CUDA Graph replay.
On Llama-3.1-8B, the resulting pipeline achieves up to
\textbf{$1.51\times$ prefill throughput} and
\textbf{$1.22\times$ Graph decoding speedup} over the corresponding FP16
baseline, while reducing Graph decoding peak memory by
\textbf{$56.34\%$}.
Relative to the matched QuaRot--Hadamard pipeline, the additional
Graph decoding latency is only $2.35\%$.
The experiments below separate quantization-design choices, local kernel
improvements, and measured whole-model performance.

\subsection{Design Trade-offs and the Low-Rank Operating Point}
\label{app:deployment_tradeoffs}

\begin{figure}[!tbp]
    \centering
    \includegraphics[width=\linewidth]{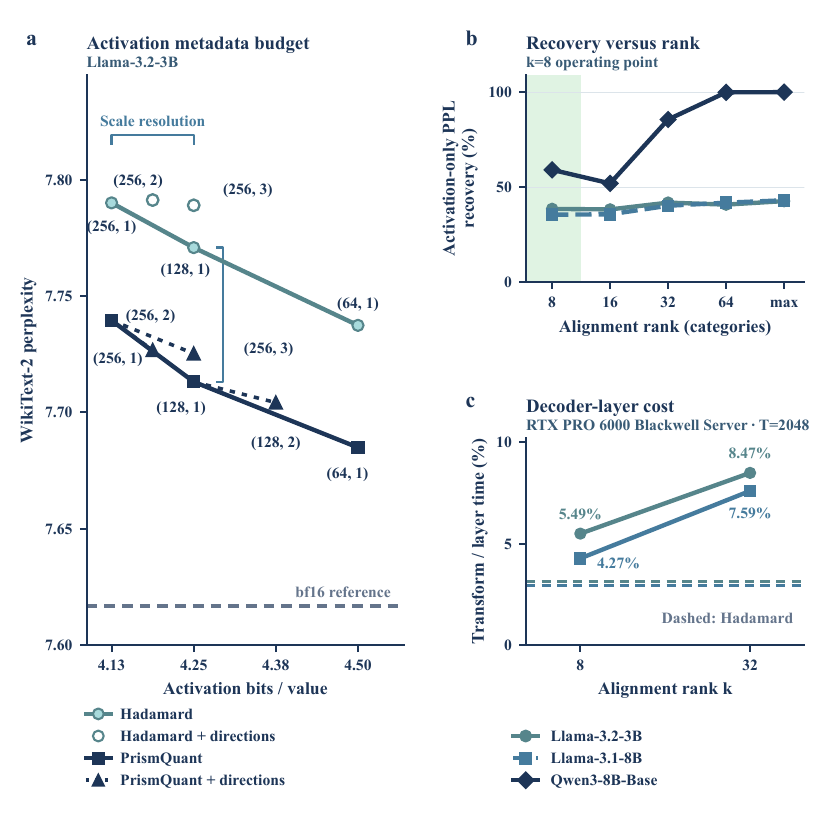}
    \caption{\textbf{Activation design and local transform cost.}
    \textbf{(a)}~Activation representation budget versus WikiText-2 PPL
    on Llama-3.2-3B; $(g,m)$ denotes group size and the number of
    represented directions.
    \textbf{(b)}~Activation-only PPL recovery versus rank.
    The highlighted $k=8$ category is the deployment operating point;
    the actual rank is capped separately at each quantization site.
    Llama results use three seeds and 64 chunks per seed; Qwen results
    use three seeds and 146 chunks, with bf16 weights and KV states.
    \textbf{(c)}~Original local cost measurements at $T=2048$ on an
    NVIDIA RTX PRO 6000 Blackwell Server Edition:
    $100t_{\mathrm{transform}}/
    (t_{\mathrm{decoder\ layer}}+t_{\mathrm{transform}})$.
    These separately benchmarked durations are not whole-model decode
    overhead; end-to-end deployment is evaluated separately.}
    \label{fig:app_prism_tradeoffs}
    \label{fig:budget_deployment} 
\end{figure}

Figure~\ref{fig:app_prism_tradeoffs} collects the two accuracy results
that motivate a compact implementation and adds the local cost of the
transform.
Panels~(a) and (b) plot the metadata-allocation study of
Table~\ref{tab:app_ablations}a and the rank study of
Table~\ref{tab:rank_ablation}: at a fixed activation budget of
$4+16(m+1)/g$ bits per value, alignment improves perplexity without
additional metadata, and $k=8$ already recovers most of the
Hadamard-to-bf16 gap on Llama, with larger ranks giving model-dependent
rather than monotone gains.
Panel~(c) measures the transform's share of decoder-layer time at
$T=2048$ in isolation; these local durations are not whole-model
overheads, which \Cref{app:deployment_results} reports separately.
We therefore adopt $k=8$ as the deployment operating point, not as a
claim that it maximizes accuracy for every model.

\begin{figure}[!t]
    \centering
    \includegraphics[width=\linewidth]{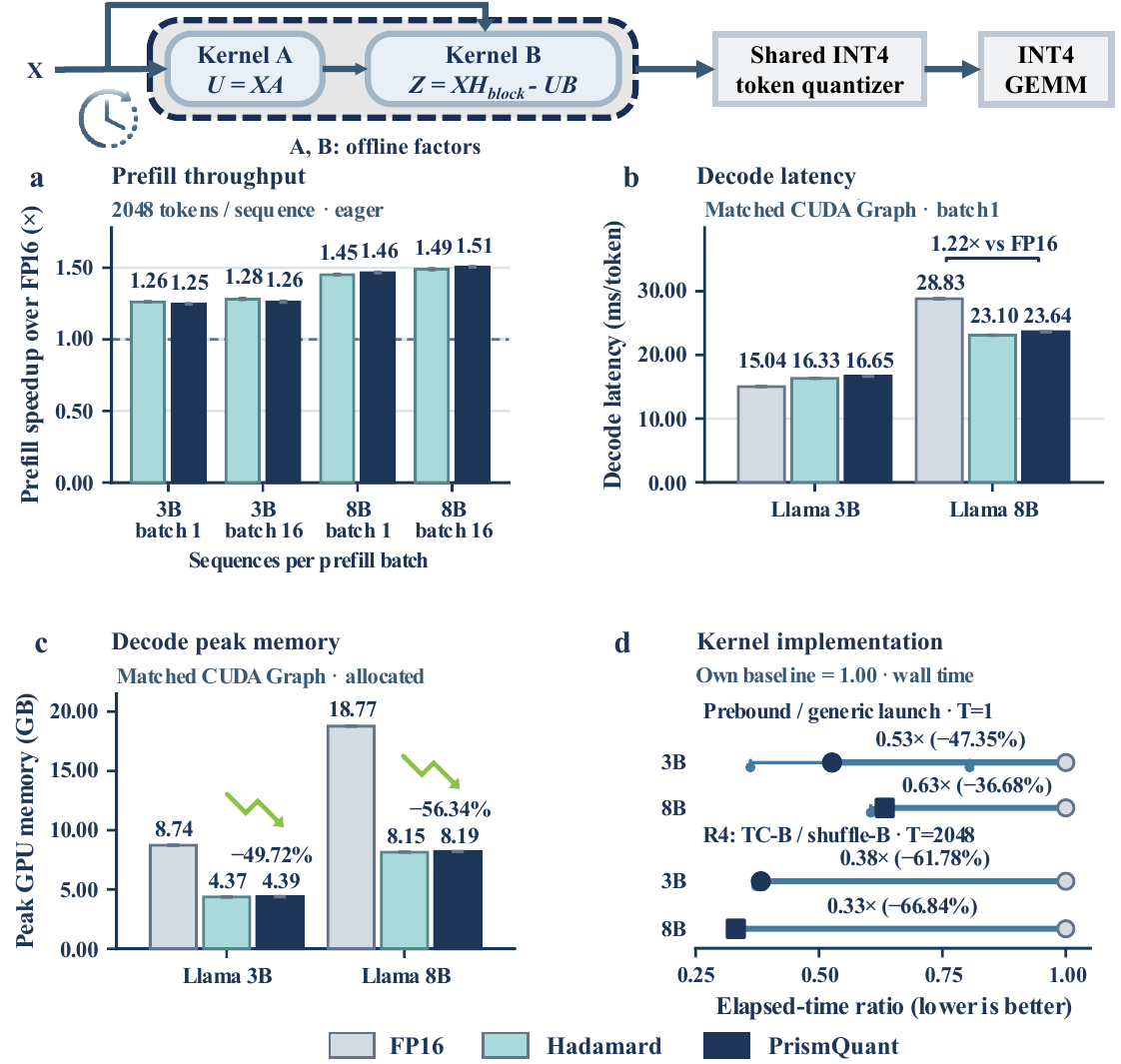}
    \caption{\textbf{Practical efficiency of the compact \pq{} implementation.}
    The upper strip shows the two-kernel $R_4$ computation followed by
    the shared INT4 quantizer and GEMM; box widths do not represent time.
    \textbf{(a)}~Eager prefill throughput relative to matched FP16,
    with 2048 input tokens per sequence.
    \textbf{(b,c)}~Matched CUDA Graph decode latency and peak allocated
    memory on A40, using batch size 1 and a 2048-token prefix.
    \textbf{(d)}~Local implementation ablations: pre-bound versus
    generic launches at $T=1$, and complete $R_4$ paths using
    Tensor Core versus shuffle-based Kernel B at $T=2048$.
    Each ablation has its own reference; these are not whole-model
    speedups.
    Timing centers are pooled medians; ablation centers are medians
    of paired-session ratios.
    Whiskers show session ranges, not confidence intervals.
    All performance models use random weights.}
    \label{fig:app_prism_efficiency}
\end{figure}

\subsection{Compact Kernels and Runtime Integration}
\label{app:deployment_kernels}

\paragraph{Two-kernel rotation.}
We reuse the existing CUTLASS INT4 matrix multiplication and replace
only the online $R_4$ transform preceding the MLP down-projection input
quantizer.
For a row-batched activation $X\in\mathbb{R}^{T\times d}$, the exported
Householder/compact-WY factors give the online computation
\begin{equation}
    U=XA,
    \qquad
    Z=XH_{\mathrm{blk}}-UB,
    \qquad
    A\in\mathbb{R}^{d\times k},\quad
    B\in\mathbb{R}^{k\times d}.
    \label{eq:app_r4_kernels}
\end{equation}
Here $H_{\mathrm{blk}}$ applies normalized Hadamard transforms to
128-channel blocks, and the offline construction absorbs the relevant
signed permutation into the factors and, for a compensated model,
the corresponding weights.
No dense $d\times d$ rotation is materialized online.
The projection and correction each require $O(Tdk)$ arithmetic;
$k$ controls the correction rank, not the number of model channels.

Kernel A evaluates $U=XA$ on Tensor Cores, with separate FP32
accumulators for the FP16 high/low factor terms.
Kernel B combines the block-Hadamard transform and low-rank correction
within each token-tile/channel-block program.
On A40, the 128-channel Hadamard is implemented using Tensor Core
matrix multiplication, and the correction retains the three leading
high/low cross-products.
This avoids writing a full-width Hadamard intermediate between these
operations; the two kernels communicate through a compact FP32 partial
buffer.
Our integration writes $Z$ in FP16 and then invokes the shared QuaRot quantizer and INT4 GEMM\citep{ashkboos2024quarot}.
It does not replace their interface with the native group-asymmetric
packing epilogue used in the separate transform microbenchmark.

\paragraph{Reducing dispatch and allocation overhead.}
Compiled kernel calls are pre-bound to avoid repeated Python/Triton
argument binding.
Temporary projection buffers and the fixed Hadamard matrix are reused
across sequential layers.
Common cache metadata is prepared once per decode step, and a
current-stream-compatible backend enables complete one-step CUDA Graph
capture for all three methods.
Replay updates positions and KV metadata while preserving the growing
causal context; it does not repeatedly evaluate a fixed cache state.
These shared runtime adaptations are applied consistently to the
FP16, Hadamard, and \pq{} paths wherever applicable.

\subsection{Benchmark Protocol}
\label{app:deployment_protocol}

We benchmark Llama-3.2-3B and Llama-3.1-8B on one NVIDIA A40
($\mathrm{sm}_{86}$), using PyTorch 2.4.1, CUDA 12.4, Triton 3.0.0,
and the QuaRot-derived integer pipeline.
The two quantized methods share packed INT4 weights, quantizer,
GEMM, attention, and KV-cache implementations; their common random
states are verified by tensor hashes.
All performance models use random weights, making this a
\emph{kernel-swap performance benchmark}, not a model-quality evaluation.
The floating-point performance reference is FP16.

Prefill processes 2048 input tokens per sequence at batch sizes 1 and 16,
with only the final-position vocabulary logits computed for all methods.
Decode uses batch size 1 after a 2048-token prefix, executes 128 causal
steps, and excludes the first 8 from steady-state timing.
Each condition has 10 warmup runs and 50 measured runs in each of three
balanced sessions; tables report pooled timing medians.
Continuous eager timing synchronizes around the measured sequence,
not between its individual decode steps.
Both decode modes in Table~\ref{tab:app_deployment} use the matched
current-stream backend; prefill uses the separately recorded eager
prefill cohort.
No speedup is computed across unmatched backends or execution modes.

Graph capture and compilation are excluded from steady-state timing,
while replay metadata updates are included.
The timed Graph layout uses one preallocated 2176-token page;
separate page-64 tests verify actual page-boundary behavior.
All six matched Graph paths pass the growing-cache replay checks.
Peak allocated memory is measured in independent processes and
reported in decimal GB, using the maximum of the three session peaks.

\begin{table}[!tbp]
\centering
\captionsetup{font=small,skip=4pt}
\caption{\textbf{End-to-end performance on a single NVIDIA A40.}
Prefill reports input tokens/s ($\uparrow$), and decode reports
ms/token ($\downarrow$).
Both decode columns use the matched current-stream backend;
Graph speedup and peak memory use CUDA Graph execution.
Speedup is relative to the FP16 row of the same model and mode.
All rows use random weights; bold marks the best quantized entry
per model and metric.}
\label{tab:app_deployment}

\begingroup
\fontsize{8pt}{9.5pt}\selectfont
\setlength{\tabcolsep}{2pt}
\renewcommand{\arraystretch}{1.12}
\setlength{\extrarowheight}{0.3pt}
\setlength{\aboverulesep}{2pt}
\setlength{\belowrulesep}{2pt}
\arrayrulecolor{black}

\begin{tabularx}{\linewidth}{l*{6}{>{\raggedleft\arraybackslash}X}}
\toprule
& \multicolumn{2}{c}{\textbf{Prefill} $\uparrow$}
& \multicolumn{2}{c}{\textbf{Decode} $\downarrow$}
& \multicolumn{2}{c}{\textbf{Graph deployment}} \\
\cmidrule(lr){2-3}
\cmidrule(lr){4-5}
\cmidrule(lr){6-7}
\textbf{Method}
& B1 & B16
& Eager & Graph
& \shortstack[r]{Peak GB\\$\downarrow$}
& \shortstack[r]{Speedup\\$\uparrow$} \\
\midrule

\rowcolor{PQTRef}
\textcolor{PQTInk}{\textbf{Llama-3.2-3B} (FP16)}
& 10,163.48 & 13,985.13
& 21.86 & 15.04
& 8.74 & $1.00\times$ \\

QuaRot--Hadamard
& \pqtdepbest{12,819.74} & \pqtdepbest{17,894.66}
& \pqtdepbest{42.83} & \pqtdepbest{16.33}
& \pqtdepbest{4.37} & $\mathbf{0.92}\times$ \\

\rowcolor{PQTOurs}
\pqtdepours{} ($k=8$)
& 12,674.40 & 17,630.27
& 43.08 & 16.65
& 4.39 & $0.90\times$ \\

\midrule

\rowcolor{PQTRef}
\textcolor{PQTInk}{\textbf{Llama-3.1-8B} (FP16)}
& 5,241.70 & 6,418.43
& 30.22 & 28.83
& 18.77 & $1.00\times$ \\

QuaRot--Hadamard
& 7,607.40 & 9,556.00
& 52.31 & \pqtdepbest{23.10}
& \pqtdepbest{8.15} & $\mathbf{1.25}\times$ \\

\rowcolor{PQTOurs}
\pqtdepours{} ($k=8$)
& \pqtdepbest{7,677.01} & \pqtdepbest{9,661.75}
& \pqtdepbest{49.99} & 23.64
& 8.19 & $1.22\times$ \\

\bottomrule
\end{tabularx}
\endgroup
\end{table}

\subsubsection{End-to-End Gains and Memory Efficiency}
\label{app:deployment_results}

\paragraph{Preserving prefill throughput.}
Figure~\ref{fig:app_prism_efficiency}(a) and
Table~\ref{tab:app_deployment} show that the additional rotation
structure has little impact on the throughput of the integer pipeline.
On 3B, \pq{} retains $98.87\%$ and $98.52\%$ of Hadamard's prefill
throughput at batch sizes 1 and 16.
On 8B, it reaches $100.92\%$ and $101.11\%$, respectively.
Thus, all four prefill conditions remain within $1.50\%$ of Hadamard,
while \pq{} achieves $1.25$--$1.51\times$ the matched FP16 throughput.
The small 8B batch-1 advantage is interpreted as near parity rather
than a robust throughput improvement.

\paragraph{Efficient decoding with small incremental cost.}
In the matched 8B eager comparison, \pq{} reduces decode latency from
$52.31$ to $49.99$ ms/token, a $4.43\%$ reduction.
Graph replay further reduces \pq{}'s latency from $43.08$ to
$16.65$ ms/token on 3B and from $49.99$ to $23.64$ ms/token on 8B,
corresponding to $2.59\times$ and $2.11\times$ improvements over its
own matched eager execution.
These gains measure the execution-mode change, not a rotation-only
speedup over Hadamard.
When both methods use Graph replay, \pq{} incurs only $1.96\%$ and
$2.35\%$ additional latency on 3B and 8B.
On 8B, it also achieves $1.22\times$ faster Graph decoding than FP16;
on 3B, the $16.65$ ms/token result remains above the FP16 reference
of $15.04$ ms/token.
Together, these results show that the more structured rotation can
retain competitive end-to-end performance at a small incremental cost.

\paragraph{Substantial memory savings.}
Under the same Graph mode, \pq{} reduces peak allocated memory from
$8.74$ to $4.39$ GB on 3B and from $18.77$ to $8.19$ GB on 8B,
corresponding to savings of $49.72\%$ and $56.34\%$.
Its resident high/low factors occupy $96d$ bytes per layer:
$64d$ for the rank-padded projection factors and $32d$ for the
correction factors.
Across all layers, these factors require $22.02$ MB on 3B and
$44.04$ MB on 8B, plus one shared 32 KiB Hadamard matrix and
shape-dependent scratch storage.
This compact representation explains why \pq{} remains close to
Hadamard's memory footprint despite its richer transform.
The reported peaks are inference memory, not serialized weight sizes.

\subsection{Kernel-Level Evidence and Measurement Scope}
\label{app:deployment_kernel_results}

Table~\ref{tab:app_frontend} isolates the directly timed
\emph{rotation-plus-quantization frontend}, with the same QuaRot
quantizer after either transform.
On 8B, the optimized \pq{} frontend is faster at all three tested
shapes, with paired speedups of $1.12$--$1.73\times$.
On 3B, it improves the single-token frontend by $1.25\times$, while
larger token batches incur extra local cost.
These shape-dependent measurements are consistent with the small
3B prefill throughput reduction observed end to end; a local speedup
or slowdown should not be equated with the same whole-model change.

\begin{table}[!tbp]
\centering
\captionsetup{font=small,skip=4pt}
\caption{\textbf{Rotation-plus-quantization frontend speedup over QuaRot.}
Each value is the reciprocal of the median of three paired-session
\pq{}/Hadamard elapsed-time ratios.
Values above $1.00\times$ indicate faster execution and are bold.
These are directly measured local frontend chains, not end-to-end
model speedups; $T$ is the number of input activation rows.}
\label{tab:app_frontend}

\begingroup
\fontsize{8.5pt}{10pt}\selectfont
\setlength{\tabcolsep}{5pt}
\renewcommand{\arraystretch}{1.12}
\setlength{\aboverulesep}{2pt}
\setlength{\belowrulesep}{2pt}
\arrayrulecolor{black}

\begin{tabularx}{\linewidth}{l*{3}{>{\raggedleft\arraybackslash}X}}
\toprule
\textbf{Model}
& $T=1$ & $T=2048$ & $T=32768$ \\
\midrule
Llama-3.2-3B
& $\mathbf{1.25}\times$ & $0.93\times$ & $0.85\times$ \\
\rowcolor{PQTOurs}
\textcolor{PQTInk}{\textbf{Llama-3.1-8B}}
& $\mathbf{1.73}\times$ & $\mathbf{1.17}\times$ & $\mathbf{1.12}\times$ \\
\bottomrule
\end{tabularx}
\endgroup
\end{table}

The controlled implementation ablations in
Figure~\ref{fig:app_prism_efficiency}(d) support both design choices.
At $T=1$, pre-bound launches reduce the local elapsed-time ratio to
$0.53$ on 3B and $0.63$ on 8B relative to generic launches of the
same compiled arithmetic.
At $T=2048$, replacing shuffle-based Kernel B with its Tensor Core implementation reduces the complete $R_4$ path ratio to $0.38$ and $0.33$, respectively.
The latter ratios compare full transform paths, not Kernel B alone; all other tested shapes, including unfavorable cases, are retained in the accompanying experiment records.

\paragraph{Scope of the deployment evidence.}
These experiments establish compact-transform efficiency within a
shared packed-INT4 pipeline.
The performance backend uses per-token symmetric activation
quantization and KV interfaces distinct from the paper's
native group-128 asymmetric accuracy protocol; the two experiments
do not constitute a checkpoint-level joint accuracy--speed measurement.
The recorded large-accumulator FP16-conversion failures in the shared
backend remain an explicit numerical limitation, separate from the
passed rotation and growing-cache Graph checks.
Within this scope, the results demonstrate that \pq{} preserves
low-bit prefill and memory benefits with only a small additional
Graph decoding cost.

%
%
\clearpage

\section{Local Activation Visualizations}
\label{app:local_visualizations}

\begin{figure*}[!htbp]
    \centering
    \captionsetup{font=small,skip=4pt}
    \includegraphics[width=\linewidth]{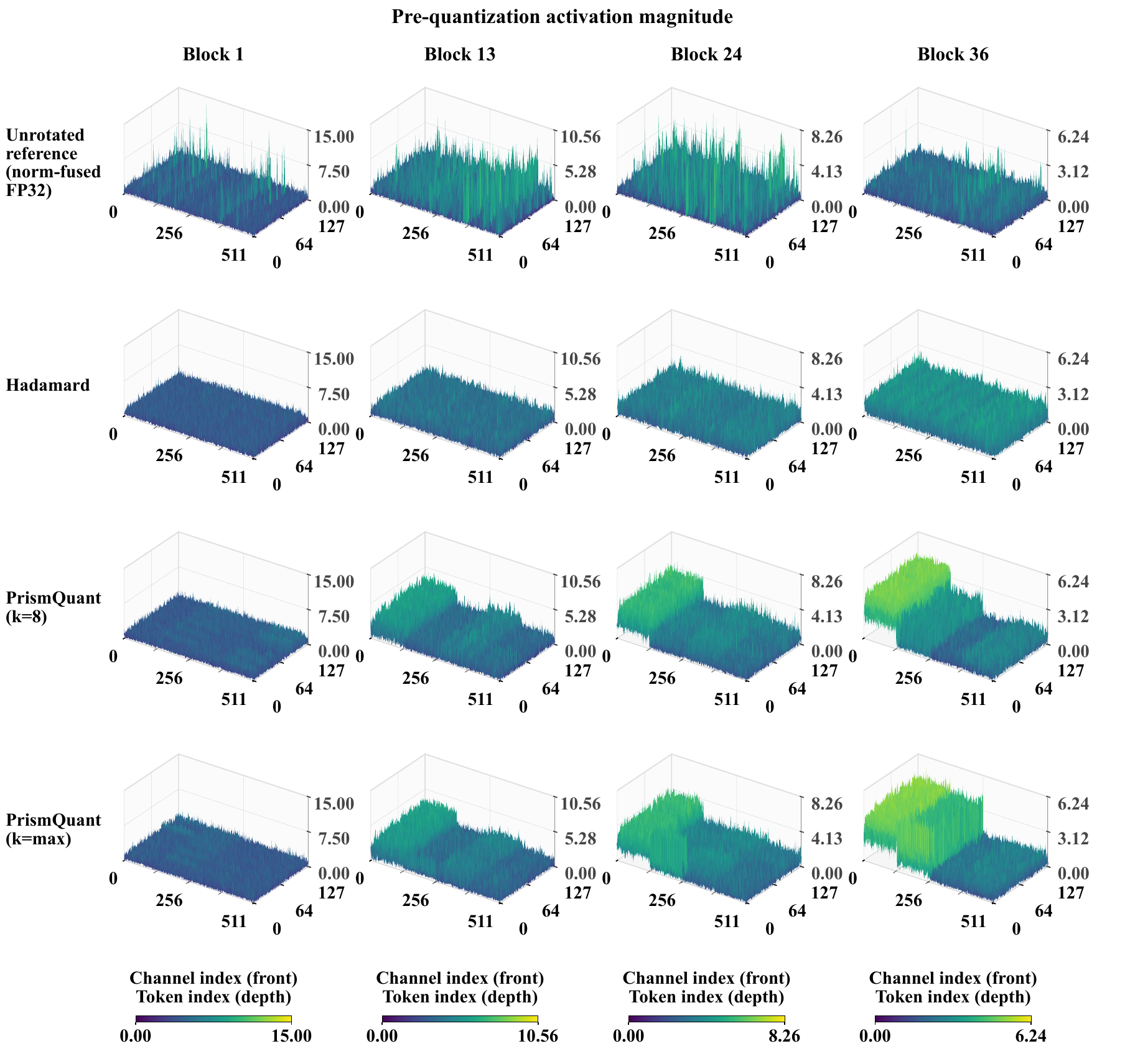}

    \caption{\textbf{Local end-to-end activation magnitudes at
    \texttt{q\_proj} on Qwen3-8B-Base.}
    Columns show Blocks 1, 13, 24, and 36; rows show the unrotated
    reference, Hadamard, \pq{} ($k=8$), and \pq{} ($k=\max$).
    The window is sample 0, tokens 0--127, and channels 0--511,
    with $g=128$.
    Heights represent absolute input activations before quantization.
    Linear heights and square-root colors share limits across all
    four rows within each column; the pale plane marks $z=0$.
    The rotated rows include upstream quantization effects.}

    \label{fig:qwen_local_qproj_e2e}
\end{figure*}

\begin{figure*}[!htbp]
    \centering
    \captionsetup{font=small,skip=4pt}
    \includegraphics[width=\linewidth]{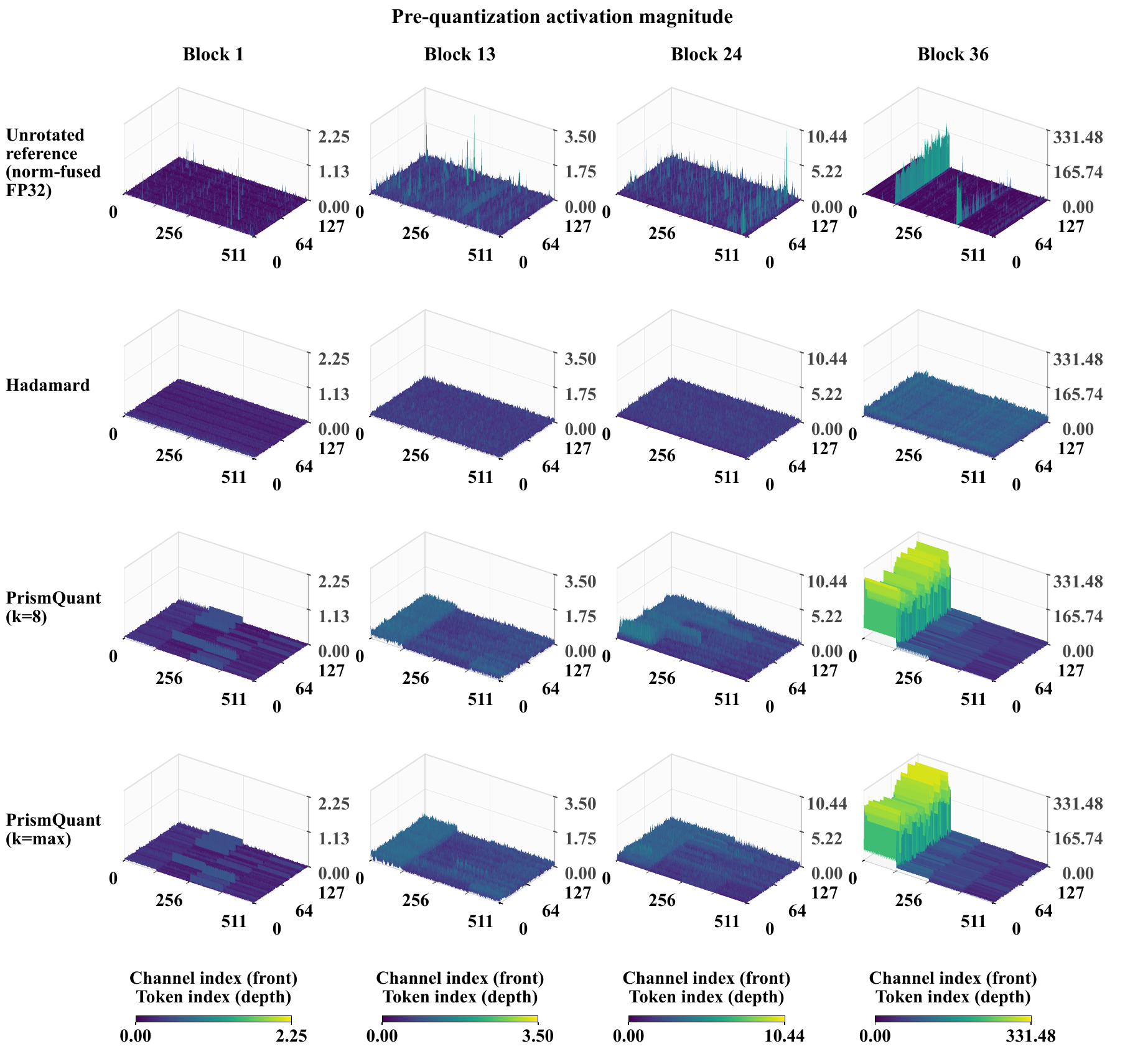}

    \caption{\textbf{Local end-to-end activation magnitudes at
    \texttt{down\_proj} on Qwen3-8B-Base.}
    The rotated rows are observed after the online $R_4$ transform
    and before activation quantization.
    Columns show Blocks 1, 13, 24, and 36; rows show the unrotated
    reference, Hadamard, \pq{} ($k=8$), and \pq{} ($k=\max$).
    We retain sample 0, tokens 0--127, and channels 0--511,
    corresponding to four groups with $g=128$.
    Heights are absolute activations, not centered residuals.
    Each column shares linear height limits and square-root color
    mapping across its four rows, with a pale $z=0$ plane.}

    \label{fig:qwen_local_downproj_e2e}
\end{figure*}

\begin{figure*}[!htbp]
    \centering
    \captionsetup{font=small,skip=4pt}
    \includegraphics[width=\linewidth]{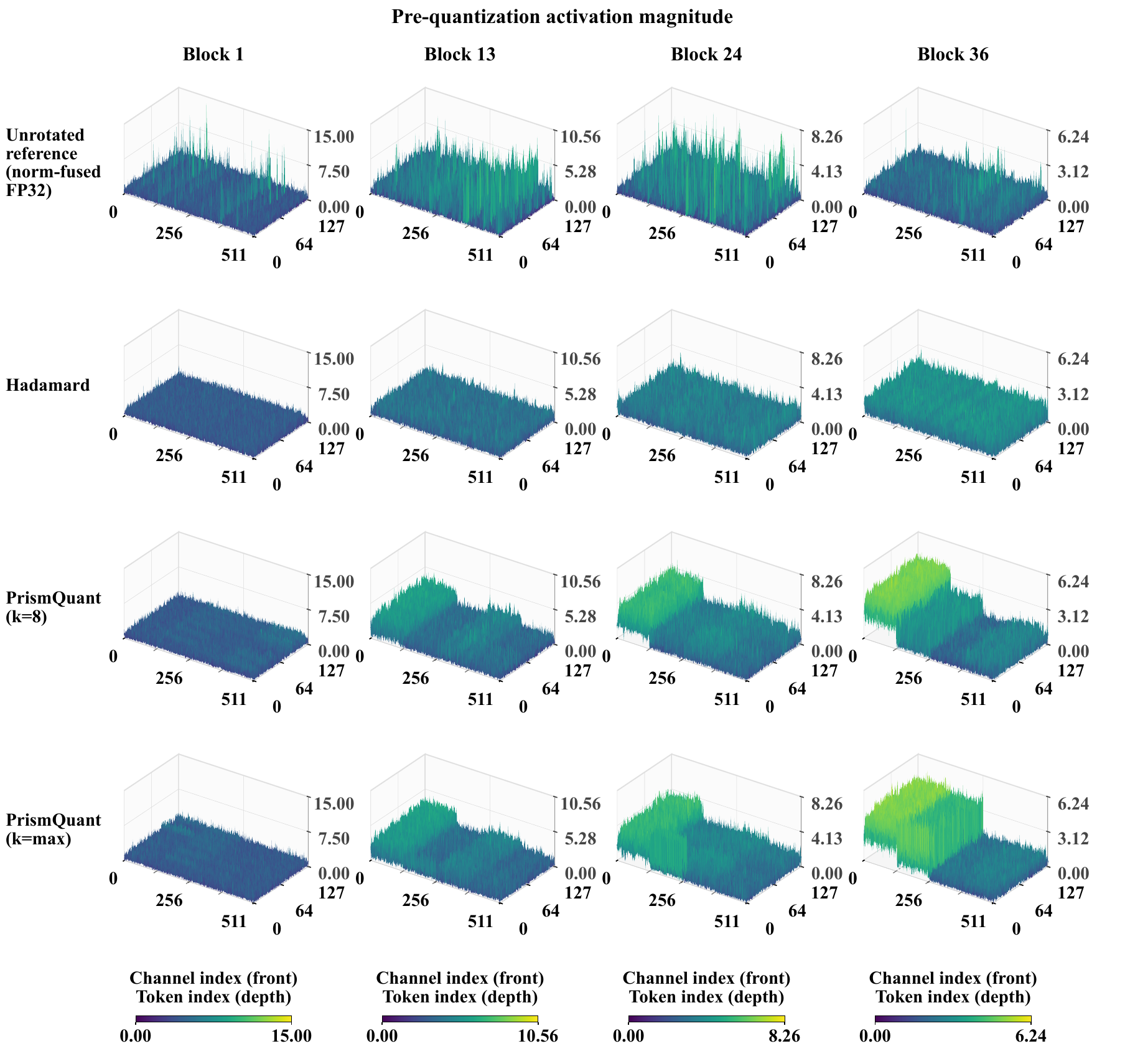}

    \caption{\textbf{Paired-local activation magnitudes at
    \texttt{q\_proj} on Qwen3-8B-Base.}
    All methods transform the same captured input tensor within
    each block, isolating the local effect of rotation.
    Columns show Blocks 1, 13, 24, and 36; rows show the unrotated
    reference, Hadamard, \pq{} ($k=8$), and \pq{} ($k=\max$).
    The displayed window contains sample 0, tokens 0--127, and
    channels 0--511, with $g=128$ and no pooling.
    Heights are linear and colors use square-root mapping,
    with identical local limits across the four rows of each
    column; the pale plane marks $z=0$.}

    \label{fig:qwen_local_qproj_paired}
\end{figure*}

To complement the quantitative alignment, range, and quantization-error
analyses, we visualize local activation magnitudes from Qwen3-8B-Base.
Figures~\ref{fig:qwen_local_qproj_e2e},
\ref{fig:qwen_local_downproj_e2e}, and
\ref{fig:qwen_local_qproj_paired} show end-to-end query-projection
inputs, end-to-end down-projection inputs, and paired-local
query-projection inputs, respectively.
Each figure contains four columns corresponding to decoder Blocks
1, 13, 24, and 36, and four rows corresponding to the unrotated
reference, Hadamard, \pq{} ($k=8$), and \pq{} ($k=\max$).
The unrotated reference is obtained from an unquantized, norm-fused
FP32 forward pass using the original bf16 checkpoint values.
The end-to-end views use the existing quantize--dequantize (QDQ)
evaluation pipeline.

All three figures use the same local window: sample 0, tokens
0--127, and channels 0--511, covering four complete quantization
groups of size $g=128$.
Every point in this window is retained without pooling or striding.
The surface height is the absolute activation value $|Y_{t,c}|$,
not a group-centered residual or a quantization reconstruction error.
Heights are linear; the blue--orange colors use a square-root mapping
to distinguish moderate amplitudes without changing their heights.
Within each column, all four rows share the same local height and
color limits; no method receives independent normalization.
The pale plane indicates $z=0$.
These fixed-window visualizations are descriptive examples rather
than aggregates over the evaluation set.


\paragraph{End-to-end query-projection inputs.}
Figure~\ref{fig:qwen_local_qproj_e2e} visualizes the residual-stream
activations entering \texttt{q\_proj}, rather than the projected
query vectors.
In the rotated models, the global $R_1$ basis is already represented
in the residual stream, and the observer records its values immediately
before input quantization without applying another rotation.
The shared-scale comparison exposes how activation magnitude is
distributed across tokens, channels, and depth during the full
quantized forward path.
Unlike the unrotated reference, the rotated rows also reflect
upstream weight and activation quantization effects.
This view therefore characterizes the resulting end-to-end activation
geometry; the paired-local comparison below isolates the effect of
the rotation itself.

\paragraph{End-to-end down-projection inputs.}
Figure~\ref{fig:qwen_local_downproj_e2e} examines the gated MLP
activations after the layer-specific online $R_4$ transform and
immediately before the \texttt{down\_proj} input quantizer.
This is the online site at which \pq{} directly aligns activation
directions with the group-constant subspace.
The local window permits inspection of four consecutive quantization
groups without channel binning.
Importantly, the method does not require uniformly smaller absolute
peaks: a large component shared within a group can be represented
by the affine offset without widening that group's range.
The raw surfaces should therefore be read as evidence of energy
redistribution, together with the quantitative group-range and
quantization-error analyses, rather than as a direct ranking of
quantization quality by peak height.

\paragraph{Paired-local query-projection inputs.}
Figure~\ref{fig:qwen_local_qproj_paired} applies the different
transforms to identical captured \texttt{q\_proj} inputs from the
unquantized reference forward pass.
Each \pq{} variant uses its deployed global $R_1$, not a separately
fitted rotation for each displayed block.
Holding the input tensor fixed separates the local basis change
from upstream quantization effects and complements the end-to-end
view in Figure~\ref{fig:qwen_local_qproj_e2e}.
Because rotation acts on the full feature dimension before the
512-channel window is selected, a change in local peak height need
not imply a change in total activation energy.
Together, the paired-local and end-to-end views distinguish direct
rotation-induced redistribution from the activation geometry
observed along the quantized forward path.

\end{document}

%% file: math_commands.tex
\usepackage{amsmath,amsfonts,bm}

\def\eqref#1{equation~\ref{#1}}

\def\1{\bm{1}}

\DeclareMathAlphabet{\mathsfit}{\encodingdefault}{\sfdefault}{m}{sl}
\SetMathAlphabet{\mathsfit}{bold}{\encodingdefault}{\sfdefault}{bx}{n}

